\documentclass{article}
\usepackage{ijcai24}

\usepackage{times}
\usepackage{soul}
\usepackage{url}
\usepackage[hidelinks]{hyperref}
\usepackage[utf8]{inputenc}
\usepackage[small]{caption}
\usepackage{graphicx}
\usepackage{amsmath}
\usepackage{amsthm}
\usepackage{booktabs}
\usepackage[ruled,linesnumbered]{algorithm2e}

\usepackage[switch]{lineno}
\usepackage{enumitem}
\usepackage{subcaption}
\usepackage{amssymb}
\usepackage{amsmath}
\usepackage{multirow}
\usepackage{dblfloatfix}
\usepackage{float}
\usepackage{algpseudocode}
\usepackage{xcolor}

\title{FedMID: A Data-Free Method for Using Intermediate Outputs as a \linebreak Defense Mechanism Against Poisoning Attacks in Federated Learning}

\author{
Sungwon Han\thanks{Equal contribution to this work.}$^1$
\and
Hyeonho Song\footnotemark[1]$^2$\and
Sungwon Park$^1$\And
Meeyoung Cha$^{4,3,1}$
\affiliations
$^1$School of Computing, KAIST, Daejeon, Republic of Korea\\
$^2$Samsung Research, Seoul, Republic of Korea\\
$^3$Data Science Group, Institute for Basic Science, Daejeon, Republic of Korea\\
$^4$Max Planck Institute for Security and Privacy, Bochum, Germany
\emails
\{lion4151, psw0416\}@kaist.ac.kr,
hyeonho.song@samsung.com,
mcha@ibs.re.kr
}

\newcommand{\suw}[1]{\textcolor{black}{#1}}

\newcommand{\cutparagraphup}{\vspace*{-0.025in}}

\begin{document}

\maketitle

\newcommand{\model}{\textsf{FedMID}}
\begin{abstract}
Federated learning combines local updates from clients to produce a global model, which is susceptible to poisoning attacks.
Most previous defense strategies relied on vectors derived from projections of local updates on a Euclidean space; however, these methods fail to accurately represent the functionality and structure of local models, resulting in inconsistent performance.
Here, we present a new paradigm to defend against poisoning attacks in federated learning using functional mappings of local models based on intermediate outputs.
Experiments show that our mechanism is robust under a broad range of computing conditions and advanced attack scenarios, enabling safer collaboration among data-sensitive participants via federated learning. \looseness=-1
\end{abstract}
\section{Introduction}
Federated learning is a decentralized machine learning approach involving multiple data-sensitive clients in producing a global optimization. 
It enables different entities in business domains to collaborate innovatively without compromising data privacy, for instance, in areas like healthcare, autonomous vehicles, and mobile applications, where data silos prevent entities from sharing information~\cite{bonawitz2019towards,park2022knowledge,voigt2017eu,ye2023personalized}.
FedAvg~\cite{mcmahan2017communication} is one of the most prominent frameworks, where the parameter updates from local clients (i.e., local updates) are averaged into a single global update and subsequently form the global model.
Unfortunately, the decentralized nature of federated learning renders the system vulnerable to poisoning attacks~\cite{baruch2019little,lyu2020threats}, where attackers disguised as local benign clients can infiltrate the system, transmit deceptive updates to impede learning, and introduce malicious knowledge into the global model~\cite{shafahi2018poison,steinhardt2017certified,park2023feddefender}. \looseness=-1

\begin{figure}
    \centering
    \begin{subfigure}{.22\textwidth}
        \centering
        \includegraphics[height=4.2cm]{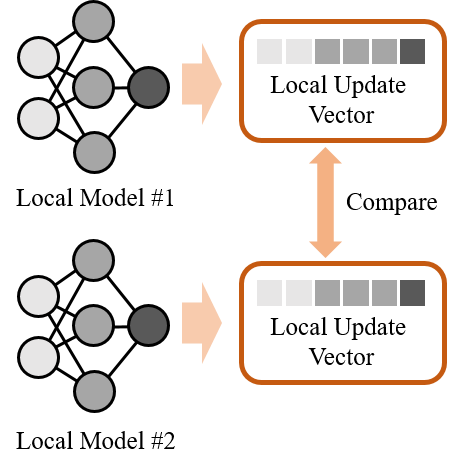}
        \caption{Parameter-based approach}
        \label{fig:motive1a}
    \end{subfigure}
    \hspace{1.3mm}
    \begin{subfigure}{.22\textwidth}
        \centering
        \includegraphics[height=4.2cm]{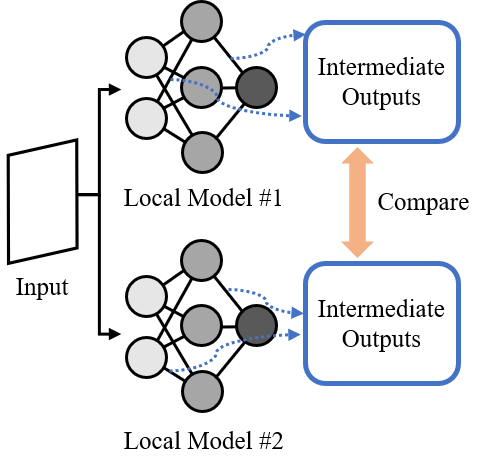}
        \caption{Proposed approach}
        \label{fig:motive1b}
    \end{subfigure}
    \caption{Illustration of the proposed approach.
    (a) Parameter-based approach measures the difference in knowledge between two models by comparing their local update vectors; (b) Our approach compares intermediate outputs of two models for the same input to measure the difference in knowledge. \looseness=-1}
    \label{fig:motive}
\end{figure}


Many defense mechanisms have concentrated on devising attack-tolerant aggregation strategies that aim to identify local updates transmitted by malicious clients, mitigate their impact, and ascertain the appropriate global update~\cite{blanchard2017machine,fung2020limitations,fu2019attack,pillutla2022robust,xie2018generalized,yin2018byzantine}.
These mechanisms, for the most part, are \textit{parameter-based approaches} in that they treat model parameter updates as vectors (i.e., local update vectors) and detect outliers by scrutinizing the distance between local update vectors within Euclidean space.
For example, outlier-resistant statistics like trimmed mean and median were used to replace conventional averaging of local updates to filter out extreme values in each coordinate~\cite{xie2018generalized,yin2018byzantine}. 
Blanchard et al. proposed Krum to identify clients with updates farthest from neighboring updates as malicious~\cite{blanchard2017machine}.
However, these approaches showed inconsistent detection performance under non-IID (i.e., non-independent, identically distributed) data settings, increased number of local training epochs, and advanced attack strategies when partial information about benign clients is known~\cite{baruch2019little,fung2020limitations,fang2020local,shejwalkar2021manipulating,wang2022defense}.

This paper identifies two significant limitations of parameter-based approaches: {functional inconsistency} and {structural inconsistency}.
\textbf{Functional inconsistency} implies that the local update vector alone cannot fully represent changes in the model's functional mapping.
Functional mapping refers to how the model transforms input data into output predictions through its learned decision boundaries, which embody the model's acquired knowledge~\cite{bhat2021distill}.
Interestingly, two models with different parameters can exhibit identical functional mappings, while deep learning models can have similar parameters yet perform different functions~\cite{yurochkin2019bayesian,wang2020federated}.
Based on these characteristics, Baruch et al. showed that poisoning attacks could be successful when generating false models with directed small changes to many parameters~\cite{baruch2019little}.
This finding suggests that the local update vector is \textit{insufficient} for accurately identifying malicious local models with different functionalities. 

\textbf{Structural inconsistency}, on the other hand, implies parameters cannot adequately represent the model's architecture and scale differences. The model can have various types of layers, and their weights and biases inherently possess different scales in updates~\cite{lee2023layer,li2021fedbn}.
Treating them as a single vector overlooks these scale disparities; consequently, parameter-based approaches can no longer maintain consistent performance as the model's size and architecture change.
We empirically confirm that both types of inconsistencies can negatively affect the prediction performance of malicious clients in a federated system.
\looseness=-1

This paper presents a new paradigm \model{} (Federated learning with Model's Intermediate output-based Defense) for defending against poisoning attacks in federated learning. Our approach directly measures the knowledge difference between benign and malicious local models.
We bring the concept of knowledge distillation~\cite{hinton2015distilling,kim2021self,han2022fedx} and compare the model's functional mapping by examining intermediate outputs (see Fig.~\ref{fig:motive1b}) as opposed to comparing model parameters, as is the case with existing methods (see Fig.~\ref{fig:motive1a}). 
The use of random synthetic datasets sampled from the standard normal distribution is one of the crucial aspects of this study. This eliminates the need to acquire local data information that can be privacy concerning or the need to use open-sourced public datasets, which can limit application diversity.
\looseness=-1

By comparing the distances in the intermediate outputs of models, it is possible to directly examine the disparities in their functional mappings, thereby effectively addressing the problem of functional inconsistency.
In addition, we address the problem of structural inconsistency by normalizing the scale of distances across all intermediate layers. Extensive experiments demonstrate that our approach consistently outperforms other parameter-based methods across a broad range of simulation settings, including non-IID data settings, different model architectures, varying numbers of local epochs, and adaptive attack scenarios, while maintaining comparable computational costs. The code and implementation details will be made available soon.

\section{Problem Formulation}
\cutparagraphup
\paragraph{Federated optimization~~} Consider a federated learning (FL) system with $N$ clients. 
Each client $i$ possesses a training dataset, denoted as $\mathcal{D}_i$ ($i\in {1, ..., N}$).
The primary goal of FL is to train a shared global model with parameters $\phi$ without directly sharing the local dataset $\mathcal{D}_i$ among the clients.
Given the loss objective $\mathcal{L}_{i}$ for each client $i$, the optimization objective for $\phi$ is \looseness=-1
\begin{equation}
    \min_{\phi} \mathcal{L}(\phi) = \min_{\phi} {\sum_{i=1}^{N} |\mathcal{D}_i| \cdot \mathcal{L}_i(\phi, \mathcal{D}_{i}) \over \sum_{i=1}^{N} |\mathcal{D}_i|}.
    \label{eq:fed_objective}
\end{equation}
FedAvg~\cite{mcmahan2017communication} is commonly used as the computing environment to optimize the goal above. 
This algorithm divides the training process into multiple iterative steps, where at the $t$-th iteration ($t\geq 0$), the central server shares the global model $\phi^{t}$ with a random subset of clients.
The chosen clients then update their local model weights, $\theta_i^t$, based on respective datasets $\mathcal{D}_i$. 
These updates, denoted as $\Delta_i^t = \theta_i^{t} - \phi^{t}$, are sent back to the central server. 
The server aggregates the received local model updates and adjusts the global model weights $\phi^{t+1}$ via an \emph{aggregation process}:
\begin{equation}
\phi^{t+1} = \phi^t + {{\sum_{i\in [1..N]} |\mathcal{D}_i| \cdot \Delta_i^t} \over \sum_{i\in [1..N]} |\mathcal{D}_i|}. \label{eq:fedavg}
\end{equation}

\cutparagraphup
\paragraph{Threat model~~} FL systems assume that participating clients are benign and that their updates can be trusted. 
As a result, the aggregation process can be compromised by malicious attacks that corrupt or tamper with local updates. 
In order to poison the global model, adversaries can either disrupt its convergence (called an ``untargeted'' attack) or embed a backdoor with a hidden trigger (called a ``targeted'' attack).
We hypothesize that adversaries can circumvent the verification process and join the system as clients. 
We mainly consider three threat scenarios of varying attack capability: (1) attackers have no access to other benign clients' information (i.e., non-omniscient attack), (2) attackers have partial access to other benign clients' information, such as the standard deviation or averaged updates from benign clients (i.e., omniscient attack), and (3) attackers have knowledge of the defense and can circumvent it (i.e., adaptive attack). We also assume that the number of malicious clients does not outnumber benign clients.
\looseness=-1

\begin{figure*}
    \centering
    \begin{subfigure}{.32\textwidth}
        \centering
        \includegraphics[height=3.4cm]{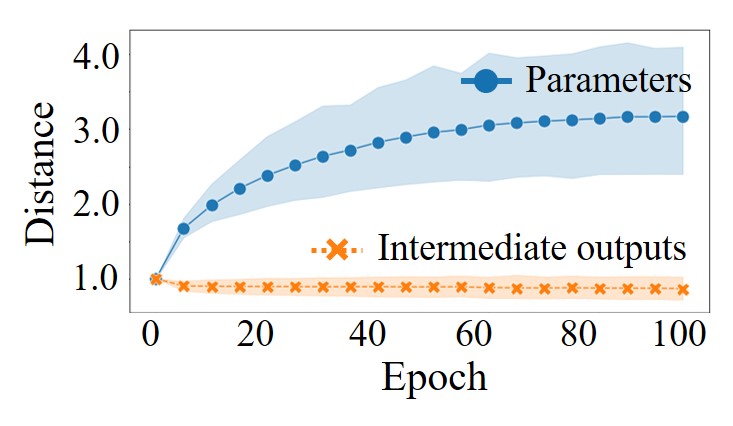}
        \caption{Distance among benign clients}
        \label{fig:motive2a}
    \end{subfigure}
    \hfill
    \begin{subfigure}{.345\textwidth}
        \centering
        \includegraphics[height=3.4cm]{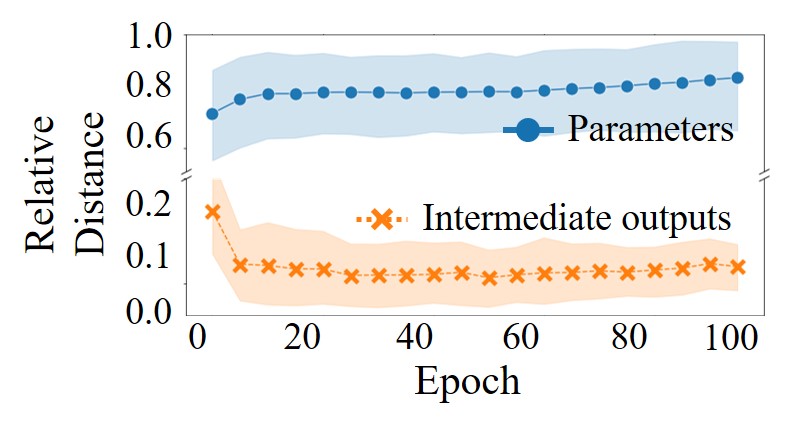}
        \caption{Relative distance ratio ($\text{dist}_{b} / \text{dist}_{b,m}$)}
        \label{fig:motive2b}
    \end{subfigure}
    \hfill
    \begin{subfigure}{.31\textwidth}
        \centering
        \includegraphics[height=3.4cm]{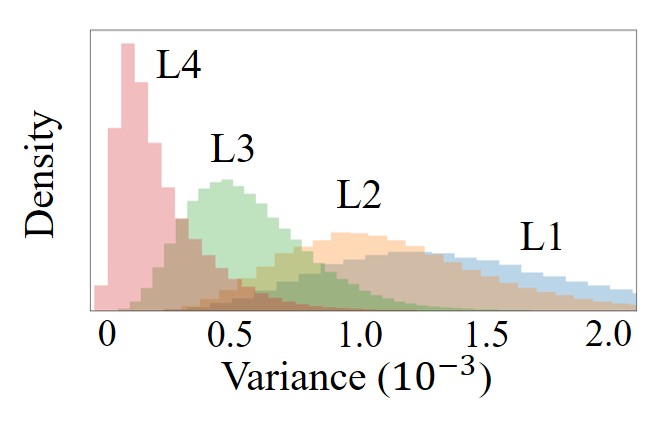}
        \caption{Histogram of variance in updates \looseness=-1}
        \label{fig:motive2c}
    \end{subfigure}
    \caption{Empirical evidence demonstrating the limitations of the parameter-based approach: (a) While parameter-based distance shows divergence as training progresses, intermediate outputs-based distance exhibits minimal changes or reductions. (b) Parameter-based distance reveals a higher ratio of the intra-distance among benign clients to the inter-distance between malicious and benign clients (i.e., $\text{dist}_{b} / \text{dist}_{b-m}$). 
    (c) The variance differs by the location of layers, with the former layers exhibiting greater variance (L1: initial intermediate blocks, L4: the final intermediate blocks).\looseness=-1}
    \vspace*{-3mm}
    \label{fig:motive2}
\end{figure*}

\cutparagraphup
\paragraph{Attack-tolerant aggregation~~}  
Several previous studies have proposed attack-tolerant aggregation techniques that can identify malicious updates during the aggregation process~\cite{fu2019attack,xie2018generalized,yin2018byzantine}.
Consider two disjoint sets of clients in $\mathcal{C}$, where $\mathcal{C}_m$ represents malicious clients and $\mathcal{C}_b$ denotes benign clients (i.e., $\mathcal{C}_m\cup \mathcal{C}_b = \mathcal{C}$). 
The goal here is to design a central aggregation function $\mathcal{A}(\cdot)$ that determines how the global model is updated: $\phi^{t+1} = \phi^t + \sum_{i\in \mathcal{C}} \mathcal{A}(i) \cdot \Delta_i^t$.
During the aggregation phase, the optimal aggregation function $\mathcal{A}^*(i)$ must screen every malicious update, as formalized by the following equation:
\looseness=-1
\begin{equation}
\mathcal{A}^*(i)={\mathbf{1}(i \in \mathcal{C}_b) \cdot |\mathcal{D}_i| \over \sum_{i \in \mathcal{C}_b}{|\mathcal{D}_i|}}. \label{eq:optimal_fedavg}
\end{equation}
Here, $\mathbf{1}(i \in \mathcal{C}_b)$ is an indicator function that takes the value of one if client $i \in \mathcal{C}_b$ and zero if client $i \notin \mathcal{C}_b$. 
To prevent attackers with larger datasets from amplifying false updates, the term $|\mathcal{D}_i|$ from Eq.~\ref{eq:optimal_fedavg} can be omitted. \looseness=-1

\section{Limitations of Parameter-based Approaches}

We review parameter-based defense approaches against poisoning attacks in federated learning, such as Krum~\cite{blanchard2017machine} and RFA~\cite{pillutla2022robust}.  
Prior works define malicious clients as local models that exhibit substantially different functionality or knowledge compared to benign clients. 
Local updates from each client's model parameters are concatenated into a single vector, and then anomalous vectors are excluded from aggregation by removing extreme values or by calculating the abnormality of the local update vector. The following discussion suggests two challenges of parameter-based approaches:

\cutparagraphup
\paragraph{Functional inconsistency} 
This limitation refers to the fact that the local update vector cannot completely capture the changes in a model's functional mapping. Here, we adopt the definition of functional mapping as the relationship between inputs and outputs~\cite{bhat2021distill}. According to this definition, two models exhibiting the same functional mapping should produce identical outputs when given identical inputs.
While local updates indicate the magnitude of change in each parameter, they do not reveal how these parameters interact. 
This could result in the misclassification of benign clients as malicious users due to permutation invariance, a situation in which functional mappings are similar, but their parameters differ~\cite{yurochkin2019bayesian,wang2020federated}. \looseness=-1

For example, consider a simple fully connected neural network defined as $\hat{y}=\sigma (x \cdot W_1) \cdot W_2$, where $\sigma$ is a non-linear activation function and $(W_1, W_2)$ are weight parameters. 
Biases are omitted without losing generality. 
We define an orthogonal permutation matrix $P$ that reorders rows when appended on the left. This matrix will reorder columns when appended on the right.
Note that appending any permutation matrix $P$ to each weight, such as $(W_1 P, P^T W_2)$, has no effect on the functionality of the neural network (i.e., $\hat{y} = \sigma(x \cdot W_1 P) \cdot P^T W_2$).
This knowledge is critical for explaining why functional mappings of different parameters can be similar. \looseness=-1

We demonstrate the plausibility of the aforementioned phenomenon in federated learning systems.
Consider five clients, each with a ResNet-18 model~\cite{he2016deep} with the same CIFAR-10 dataset~\cite{krizhevsky2009learning} and initial weights.
Each client is independently trained on its dataset, with random training data order.
To examine how the Euclidean distance between local update vectors of clients changes after training, we show the relative change for each epoch based on the parameter distance after the first epoch of training (see blue line in Fig.~\ref{fig:motive2a}).
We separated the model layer-by-layer to calculate the distance changes for each layer, then plotted the mean change as a dot with the standard deviation area shaded.
Even with identical training conditions, parameters begin to diverge over time.

To compare the functional mapping of updated models, using a test dataset from CIFAR-10, we computed the pairwise distance matrix among input samples over the intermediate embedding space and observed the relational knowledge across the local models~\cite{yang2022mutual}. 
The yellow line in Fig.~\ref{fig:motive2a} depicts the results. The pairwise distance did not change significantly throughout the training process and occasionally decreased. These results indicate that the similarity of model parameters is not necessarily correlated with the functional similarity of models. \looseness=-1

To understand the effect of diverging parameters, we repeated the experiment while assuming the presence of malicious clients.
Two out of ten (20\%) clients were assumed as malicious, performing a label-flipping attack. 
At the end of each training round, the intra-distance $\text{dist}_b$ among benign clients was compared to the inter-distance $\text{dist}_{b,m}$ between malicious and benign clients (i.e., $\text{dist}_{b} / \text{dist}_{b,m}$).
Here, a higher distance ratio indicates that benign clients are placed relatively far apart, making it difficult to differentiate them from malicious clients; a lower ratio indicates the opposite. Fig.~\ref{fig:motive2b} shows the experiment result. The distance ratio based on intermediate outputs is far smaller than that of parameter-based approaches. The parameter-based method shows an increasing trend in the distance ratio as the epochs progress, suggesting a negative impact on identifying adversaries. \looseness=-1

\cutparagraphup
\paragraph{Structural inconsistency} This limitation refers to the lack of structural information in parameter-based approaches.
Let us take ResNet-18 as an example, which contains multiple modules such as batch normalization layers, convolution layers, and fully connected layers. 
These modules serve distinct functions, leading to different ways parameters are connected. Consequently, the parameters in these modules can exhibit varying scales and variances in local updates when measured across different clients~\cite{lee2023layer,li2021fedbn}. 
These scale differences would be ignored if all model parameters were combined into a single vector and model similarity is measured based on Euclidean space. 
Significant scale differences within a particular module among client models can overshadow or overlook discrepancies in other modules, leading to biased detection of malicious clients.
A model-specific design is required to address such a bias, such as detecting malicious updates either layer by layer or module by module within the model. 

We conducted an experiment to examine the scale difference in updates contributed by model structures. 
We tested with a FedAvg system with 20 clients and a heterogeneous local data setting with non-IID data. 
After training had progressed to 100 communication rounds, we collected the local updates sent by clients and visualized the scale differences according to the location of layers in Fig.~\ref{fig:motive2c}.
Compared to the final intermediate blocks (L4 in the figure), see significant scale changes in different layers (as in L1).

\section{Main Approach: FedMID}
\paragraph{Overview} 
We present a new defense mechanism that looks beyond parameters by directly utilizing the functional mapping of local models via their intermediate outputs.
Our method, called \model{}, focuses on the data aggregation phase where the central server receives updates of local models and uses a synthetic dataset drawn from a
standard normal distribution. 
It compares the distance matrices of input samples over each model's intermediate embedding space to assess functional differences between models, a process that can also be interpreted as a comparison of relational knowledge.
Local models with low similarity in distance matrices to other clients are considered malicious, and their contribution on the aggregation process is reduced.
We now introduce the method in detail.\looseness=-1

\cutparagraphup
\paragraph{(Step 1) Obtain intermediate outputs from synthetic data}
We employ a synthetic dataset drawn from a standard normal distribution to obtain the intermediate outputs of models without raising privacy concerns.
However, this synthetic dataset has a significantly different distribution from the input data distribution learned by the model, and thus, it fails to adequately represent the activation distribution of the hidden layers within the model.
This can lead to concerns that these synthetic input-output pairs may not accurately reflect the model's functionality, hindering \model{} from making precise comparisons of the learned knowledge among different models and identifying malicious updates.
\looseness=-1
%

We can formulate the aforementioned problem as follows. Let us denote a dataset sampled from the original data as $\mathcal{D}_{B}$ and the random synthetic input as $\mathcal{D}_{G}$.
During the inference phase, the batch normalization layer uses the running statistics ($\mathbf{\mu}_B, \mathbf{\sigma}_B$) stored from training.
We divide the batch normalization layer into (\textit{i}) a normalization stage and (\textit{ii}) a scaling/shifting stage, denoting the outcomes of each stage as $\mathbf{\Tilde{z}}$ and $\mathbf{\hat{z}}$, respectively. 
Then the normalized activation $\mathbf{\Tilde{z}}$ of $\mathcal{D}_{B}$ in the batch normalization layer becomes $P(\mathbf{\Tilde{z}}|\mathcal{D}_{B}, \mathbf{\mu}_B, \mathbf{\sigma}_B) \approx \mathcal{N}(\mathbf{0}, \mathbf{1})$.
However, due to covariate shift, the normalized activation of $\mathcal{D}_{G}$ with the same running statistics ($\mathbf{\mu}_B, \mathbf{\sigma}_B$) becomes distinct from $\mathcal{N}(\mathbf{0}, \mathbf{1})$. This phenomenon results in a difference in intermediate outputs after the $\mathbf{\hat{z}}$ stage: $P(\mathbf{\hat{z}}|\mathcal{D}_{B}, \mathbf{\mu}_B, \mathbf{\sigma}_B, \mathbf{\gamma}, \mathbf{\beta}) \neq P(\mathbf{\hat{z}}|\mathcal{D}_{G}, \mathbf{\mu}_B, \mathbf{\sigma}_B, \mathbf{\gamma}, \mathbf{\beta})$, where 
$\mathbf{\gamma}$ and $\mathbf{\beta}$ refer to the shifting and scaling parameters. \looseness=-1

We address the aforementioned shift by using the current batch statistics ($\mathbf{\mu}_G, \mathbf{\sigma}_G$) for the synthetic input $\mathcal{D}_{G}$ instead of using the mean statistics ($\mathbf{\mu}_B, \mathbf{\sigma}_B$) for normalization. 
This design choice is similar to the treatment used in a recent knowledge distillation study~\cite{datafreeKD2022nips}. 
Even when a batch with a different distribution is used as input, $P(\mathbf{\Tilde{z}} | \mathcal{D}_G, \mathbf{\mu}_G, \mathbf{\sigma}_G) = P(\mathbf{\Tilde{z}} | \mathcal{D}_B, \mathbf{\mu}_B, \mathbf{\sigma}_B) \approx \mathcal{N}(\mathbf{0}, \mathbf{1})$ holds because they are both normalized with their own batch statistics. Subsequently, using the shared shifting and scaling parameters $\mathbf{\gamma}$ and $\mathbf{\beta}$, the following also holds: $P(\mathbf{\hat{z}}|\mathcal{D}_{B}, \mathbf{\mu}_B, \mathbf{\sigma}_B, \mathbf{\gamma}, \mathbf{\beta}) \approx P(\mathbf{\hat{z}}|\mathcal{D}_{G}, \mathbf{\mu}_G, \mathbf{\sigma}_G, \mathbf{\gamma}, \mathbf{\beta})$.

This process reduces the distribution shift in intermediate outputs that are propagated after the batch normalization layer, allowing the distribution of synthetic dataset's intermediate outputs to become more similar to that from the actual dataset used during training. As a result of this alignment, the synthetic input-output pairs can more accurately represent the model's functionality and provide a closer approximation to the behavior observed with real data.
Given the synthetic dataset and models from clients, we convert each model's batch normalization layer into a training mode to use the current batch statistics. The same input from the dataset is then passed through the intermediate layers of each model to generate intermediate outputs.
\looseness=-1

\cutparagraphup
\paragraph{(Step 2) Measure deviance in functional mapping 
} 
We next compare the functional mappings among models via intermediate outputs.
According to prior studies~\cite{yurochkin2019bayesian,wang2020federated}, \textit{parameter permutations} can occur within the model layers during training in the federated system. This problem suggests the need for a distance measure that is resilient to topological changes in the embeddings.
Motivated by the concept of relational knowledge distillation~\cite{yang2022mutual}, we extract a distance matrix between samples in the embedding space and consider them a form of structural knowledge, which we use to calculate the knowledge difference between models. 
Given a sampled synthetic input dataset $\mathcal{D}_G$, for each input $\mathbf{x}_k$ (i.e., $\mathcal{D}_G={\{\mathbf{x}_1, ..., \mathbf{x}_M }$\}), we denote the intermediate embedding vector extracted from the $l$-th layer of the model as $\mathbf{z}^{(l)}_k$\footnote{We also utilize the logit vector from the last layer as an embedding to prevent attacks that solely alter the last layer.\looseness=-1}.
In this case, the distance matrix for dataset $\mathcal{D}_G$ that can be obtained from a single $l$-th layer of each model $i$ is calculated as follows: \looseness=-1
\begin{equation}
    S^{(l)}_i [k_1, k_2] = || \mathbf{z}^{(l)}_{k_1} - \mathbf{z}^{(l)}_{k_2} ||_2,
\end{equation}
where $k_1$ and $k_2$ represent the indices of row and column in the distance matrix, respectively.
Then, the difference in functional mapping of $l$-th layer in two models $i, j$ is defined as a mean absolute difference between two distance matrices: \looseness=-1
\begin{equation}
    \text{Dist}^{(l)}(i, j) = {1 \over M^2} \sum_{k_1=1}^M \sum_{k_2=1}^M  |S^{(l)}_i [k_1, k_2] - S^{(l)}_j [k_1, k_2]|.
\end{equation}

\cutparagraphup
\paragraph{(Step 3) Assess the normality score} 
Assuming that attacked models possess different knowledge than normal models, we consider clients with models with a larger difference in distance matrices compared to other clients to have a higher probability of being malicious.
Let the set of clients participating in the $t$-th communication round be $\mathcal{C}^t$. 
We define the anomaly score for model $i$ at the $l$-th layer as follows:
\begin{equation}
\text{Anom}^{(l)}(i) = \text{Median}(\{ \text{Dist}^{(l)}(i, j) | j \in \mathcal{C}^t \}).
\end{equation}

Based on the observation that malicious attackers' models exhibit similar parameter update patterns and behaviors~\cite{fung2020limitations}, we applied the median operation instead of the average to the distance measure between models to avoid underestimating the anomaly score when multiple attacker models are mixed in a round. 
Subsequently, we unified the scale of distances for each layer using Min-Max normalization, and we defined a single model's normality score $\mathcal{N}(i)$ by averaging and negating the normalized anomaly scores across the set of all layers $\mathcal{L}$: $\mathcal{N}(i) = - {1 \over |\mathcal{L}|} \sum_{l \in \mathcal{L}} \text{Normalize}(\text{Anom}^{(l)}(i)).$

\cutparagraphup
\paragraph{(Step 4) Attack-tolerant aggregation 
}
Finally, the local updates are aggregated with an adjusting weight $\mathcal{A}_i$ which is calculated from the normality score $\mathcal{N}(i)$. 
Since the normality scores would be set to high for likely-benign clients and low for likely-malicious clients, our approach will prevent the attackers' updates from being aggregated and ensures the inclusion of updates from benign users.
We again conduct a normalization process to convert the normality scores into weights with the range from 0 to 1, following the original literature~\cite{fung2020limitations}. 
This process includes min-max normalization and the inverse sigmoid function to accentuate the difference in normality scores while minimizing the over-penalization for benign clients with low and non-zero normality scores (Eq.~\ref{eq:lambda}). 
The function $\text{Clamp}_{0\sim 1}(x)$ denotes the clamping function to the range $\left[0,1\right]$ (i.e., $\text{max}(0,\text{min}(x,1))$), and $\mathcal{\Tilde{N}}(i)$ represents the min-max normalized scores ($\mathcal{\Tilde{N}}(i) =\text{Scale}(\mathcal{N}(i))$).
\begin{align}
    a_i = \text{Clamp}_{0\sim 1}(\ln {\mathcal{\Tilde{N}}(i) \over {1 - \mathcal{\Tilde{N}}(i)}} + 0.5).  \label{eq:lambda} 
\end{align}
In addition, based on the assumption that the majority of clients are benign, we assign a weight of 1 to the largest $\lceil \frac{|\mathcal{C}^t|}{2} \rceil$ clients, considering them to be benign.
Conversely, we assign a weight of 0 to the most malicious client, identified as client $i= \text{argmin}_i a_i$.
This straightforward approach enhances the stability of training in some scenarios where attackers are not present in the current communication round.
The proposed aggregation function $\mathcal{A}$ is defined as following equation: \looseness=-1
\begin{equation}
\mathcal{A}{(i)}={{a_i}\over \sum_{j \in \mathcal{C}^t}\mathbf{1}(a_j >0)},
\end{equation}
which is applied as $\phi^{t+1} = \phi^t + \sum_{i\in \mathcal{C}^t} \mathcal{A}(i) \cdot \Delta_i^t$ to produce the global model parameter $\phi^{t+1}$ at round $t$.

\section{Experiments}
Multiple benchmark datasets are used to evaluate the efficacy of model poisoning defenses.
We conducted performance analyses under various simulation settings, including (1) levels of non-IIDness, (2) local epochs, (3) varying backbones, (4) attacker's ratio, (5) the number of clients, (6) the number of synthetic samples, and (7) another federated algorithm, FedProx. 
\suw{For results regarding scenarios (4), (5), (7), and computational complexity, please refer to the Appendix.} \looseness=-1


\subsection{Performance Evaluation}
\paragraph{Dataset and system} 
We utilize four datasets in our experiments: CIFAR-10, CIFAR-100, SVHN, and TinyImageNet. 
CIFAR-10 and CIFAR-100~\cite{krizhevsky2009learning} are image datasets, each consisting of 60,000 data points of size 32x32. 
CIFAR-10 contains 10 classes, while CIFAR-100 comprises 100 classes, such as cats, dogs, and horses.
SVHN~\cite{netzer2011reading} includes 10 classes of digits ranging from 0 to 9, with a total of 99,289 data points.
TinyImageNet~\cite{le2015tiny} features 100,000 instances of 64x64 colored images, sampled from 200 classes. \looseness=-1


We set the total number of clients and communication rounds to 20 and 100 by default, respectively.
The data distribution for class $c$ in each client is determined by the probability $p^c_i$, which is sampled from the Dirichlet distribution $Dir(N,\beta)$.
The parameter $\beta$ controls the level of non-IIDness, with a lower $\beta$ value indicating greater non-IIDness.
We set $\beta$ to 0.5 as the default value.
ResNet-18 is used as the backbone architecture, as in previous works~\cite{han2022fedx,li2021model}.
Other details are described in the Appendix. \looseness=-1

\cutparagraphup
\paragraph{Attack details}
We consider three attack scenarios of varying difficulty.
The attackers are initially sampled before the training and remain unchanged throughout the process. 
The default value of attacker ratio ($r_a={|\mathcal{C}_m| / N}$) is set to 0.2.

\begin{table*}[t!]
\setlength{\tabcolsep}{11.5pt}
\centering
\scalebox{0.82}{
\begin{tabular}{@{}c|l|ccc|ccc|c@{}}
\toprule
\multirow{3}{*}{Defense} &
  \multicolumn{1}{c|}{\multirow{3}{*}{Settings}} &
  \multicolumn{3}{c|}{\textit{Scenario 1}} &
  \multicolumn{3}{c|}{\textit{Scenario 2}} &
  \multirow{3}{*}{Average} \\ \cmidrule(lr){3-8}
 &
  \multicolumn{1}{c|}{} &
  \multicolumn{1}{c|}{Untargeted} &
  \multicolumn{2}{c|}{Targeted} &
  \multicolumn{1}{c|}{Untargeted} &
  \multicolumn{2}{c|}{Targeted} &
   \\
       & \multicolumn{1}{c|}{} & \multicolumn{1}{c|}{ACC}  & ASR & ACC  & \multicolumn{1}{c|}{ACC}  & ASR & ACC  &     \\ \midrule
None   & FedAvg~\cite{mcmahan2017communication}                & \multicolumn{1}{c|}{9.0}  & 9.0 & 5.5  & \multicolumn{1}{c|}{{2.0}}  & 8.0 & 6.0  & 6.6 \\ \midrule
\multirow{8}{*}{Parameter} &
  Median~\cite{xie2018generalized} &
  \multicolumn{1}{c|}{10.8} &
  9.5 &
  10.8 &
  \multicolumn{1}{c|}{9.0} &
  12.0 &
  10.0 &
  10.3 \\
       & Trimmed Mean~\cite{yin2018byzantine}          & \multicolumn{1}{c|}{6.8}  & 7.3 & 6.0  & \multicolumn{1}{c|}{7.0}  & 11.0 & 7.0  & 7.5 \\
       & Multi-Krum~\cite{blanchard2017machine}            & \multicolumn{1}{c|}{4.8}  & 5.3 & 7.3  & \multicolumn{1}{c|}{8.0}  & 2.0 & 2.0  & 4.9 \\
       & FoolsGold~\cite{sun2019can}             & \multicolumn{1}{c|}{12.0} & 9.5 & 12.0 & \multicolumn{1}{c|}{6.0}  & 4.0 & 11.0 & 9.1 \\
       & ResidualBase~\cite{fu2019attack}          & \multicolumn{1}{c|}{4.5}  & 9.0 & {5.0}  & \multicolumn{1}{c|}{4.0}  & 10.0 & 9.0  & 6.9 \\
       & FLTrust~\cite{cao2021fltrust}   & \multicolumn{1}{c|}{9.8}  & 7.8 & 5.5 & \multicolumn{1}{c|}{\textbf{1.0}}  & 9.0 & 5.0 & 6.3 \\  
       & DnC~\cite{shejwalkar2021manipulating}                   & \multicolumn{1}{c|}{5.8}  & 5.0 & 6.0  & \multicolumn{1}{c|}{12.0} & 5.0 & 8.0  & 7.0 \\
       & RFA~\cite{pillutla2022robust}                   & \multicolumn{1}{c|}{5.5}  & 5.8 & 5.0  & \multicolumn{1}{c|}{11.0}  & 6.0 & 3.0  & 6.0 \\
       & Bucket~\cite{karimireddy2022byzantinerobust}                & \multicolumn{1}{c|}{4.3}  & 6.5 & 6.5  & \multicolumn{1}{c|}{5.0}  & 7.0 & 4.0  & 5.5 \\
       & FedCPA~\cite{han2023towards}& \multicolumn{1}{c|}{2.8} &2.0&  5.3 & \multicolumn{1}{c|}{10.0}  &3.0&12.0 & 5.9   \\
       \midrule
Output & FedMID  (Ours)        & \multicolumn{1}{c|}{\textbf{2.2}}  & \textbf{1.5} & \textbf{3.2}  & \multicolumn{1}{c|}{3.0}  & \textbf{1.0} & \textbf{1.0}  & \textbf{2.0} \\ \bottomrule
\end{tabular}}
\caption{Summary of performance comparisons under Scenarios 1 and 2. The average rank for each evaluation metric is reported for both untargeted and targeted attack settings. On average, \model{} performs the best across various datasets and scenarios. \looseness=-1}
\label{tab:main_exp_rank}
\end{table*}

\begin{itemize}[nosep,leftmargin=1em,labelwidth=*,align=left]
\vspace*{1mm}
\item \textit{Scenario 1 (Non-omniscient Attack)} is a simple label-flipping attack that can be executed without any information about benign clients. 
Two versions of attacks are considered; 
(i) targeted attacks, where attackers insert small markers in input instances and flip their labels to the target class in order to manipulate training and  
(ii) untargeted attacks, where attackers train their local models with randomly flipped labels in order to degrade global model performance. 
The pollution ratio ($\gamma_p$) is the fraction of corrupted input-target pairs in the attacker's dataset.

\vspace*{1mm}
\item \textit{Scenario 2 (Omniscient Attack)} is an advanced attack strategy where attackers have partial information about benign clients, such as LIE~\cite{baruch2019little}. It also includes targeted and untargeted versions of label-flipping attacks. 
These attacks calibrate the malicious update using the statistics of benign updates.
\looseness=-1

\vspace*{1mm}
\item \textit{Scenario 3 (Adaptive Attack)} is the most advanced form where attackers are aware of our defense method and attempt to bypass them.
They regularize intermediate outputs from their models to resemble those from the global model. This is realized by attackers training their models with the objective $L=L_{\text{adv}} + L_{\text{reg}}$, where $L_{\text{adv}}$ is the objective for malicious attacks, as the one used in \textit{Scenario 1}, and $L_{reg}$ is a regularization term defined as follows: 
\begin{equation}
L_{\text{reg}}= {1 \over |\mathcal{D}_G|} \sum_{\mathbf{x}_k \in \mathcal{D}_G}||\mathbf{z}^{(l)}_{k,b}-\mathbf{z}^{(l)}_{k,a}||_2. \label{eq:adaptive_attack}
\end{equation}
In this definition, $\mathbf{z}^{(l)}_{k,b}$ is the $l$-th intermediate output of input $\mathbf{x}_k$ from the global model, and $\mathbf{z}^{(l)}_{k,a}$ is from the malicious model. Please refer to the Appendix for details.
\end{itemize}

\cutparagraphup
\paragraph{Evaluation}
We present the top-1 classification accuracy of the global model over the test dataset. 
To ensure a stable evaluation, we calculate the average accuracy across the final 10 epochs. 
For targeted attacks, we also report the attack success rate (ASR) for the last 10 epochs.


We consider eleven baselines: (1) FedAvg~\cite{mcmahan2017communication}, which employs an averaging approach without considering potential attacks; (2) Median~\cite{xie2018generalized}; (3) Trimmed mean~\cite{yin2018byzantine}; (4) ResidualBase~\cite{fu2019attack}; (5) Bucket~\cite{karimireddy2020byzantine}; (6) Multi-Krum~\cite{blanchard2017machine}; (7) FoolsGold~\cite{fung2020limitations}; 
(8) FLTrust~\cite{cao2021fltrust};
(9) DnC~\cite{shejwalkar2021manipulating}; (10) RFA~\cite{pillutla2022robust}; and (11) FedCPA~\cite{han2023towards}. Strategies (2--5) are coordinate-wise aggregation methods, whereas strategies (6--11) are designed to identify and filter out malicious updates during the aggregation. The Appendix provides specifics for each baseline. \looseness=-1

\cutparagraphup
\paragraph{Results}
Table~\ref{tab:main_exp_rank} summarizes the experimental results for \textit{Scenarios 1 and 2}.
We report the average rank for accuracy (ACC) and attack success rates (ASR) under different attack settings with four benchmark datasets. Please see the Appendix for full results, which reports that our approach shows the best ACC and ASR in both untargeted and targeted attacks.
Even under more advanced attack settings (\textit{Scenario 2}), \model{} is comparable to or outperforms recent baselines. \suw{Note that our model showed only a slight difference in ACC compared to FedAvg in the untargeted attack setting (see Table~\ref{tab:appendix_LIE} in the Appendix). This is because untargeted attacks usually require substantial weight updates to impair the model, whereas the weight update constraint in Scenario-2 prevents the attacker from significantly harming the global model. Therefore, removing malicious attackers did not significantly impact performance, resulting in similar outcomes for FLTrust, FedAvg, and our model. Conversely, models that mistakenly removed benign clients experienced a substantial decrease in performance.}
In addition, according to Table~\ref{tab:scenario3}, the performance degradation of the model remains minimal even when attackers are aware of the defense methods and attempt to bypass them (\textit{Scenario 3}). 
Despite a decrease in performance, our method continues to have the lowest ASR compared to RFA in the targeted attack setting.
\suw{We chose RFA for comparison because it was the best baseline when rankings were aggregated in Scenario-1. The results for other baselines are in Tables~\ref{tab:appendix_main_unt_full}-\ref{tab:appendix_tar_acc} in the Appendix.}
\looseness=-1

\begin{table}[!t]
\setlength{\tabcolsep}{3pt}
\centering
\scalebox{0.73}{\begin{tabular}{l|ccc}
\toprule
\multicolumn{1}{c|}{CIFAR10} &
  \multicolumn{1}{c}{\begin{tabular}[c]{@{}c@{}}Untargeted\\ ACC ($\uparrow$)\end{tabular}} &
  \multicolumn{1}{c}{\begin{tabular}[c]{@{}c@{}}Targeted \\ ASR ($\downarrow$)\end{tabular}} &
  \multicolumn{1}{c}{\begin{tabular}[c]{@{}c@{}}Targeted\\ ACC ($\uparrow$)\end{tabular}} \\ \midrule
FedMID & 74.8/48.1/93.0/36.7 & 11.8/0.6/20.9/0.6  & 74.3/48.2/93.1/36.1 \\
+ adaptive attack  & 70.5/48.2/93.3/34.4 & 13.6/0.7/21.0/0.6  & 74.0/49.2/93.1/33.2 \\ 
FedCPA in Scenario 1 & 74.9/42.9/93.2/36.8 & 18.5/0.6/21.0/0.7 & 68.7/39.8/93.0/38.1 \\ \bottomrule
\end{tabular}}
\caption{Performance evaluation under Scenario 3. $\uparrow$ shows that a higher value is preferable, and vice versa.
Despite the adaptive attack, \model{} shows comparable performance. \suw{(Order: CIFAR10 / CIFAR100 / SVHN / TinyImageNet)} \looseness=-1}
\label{tab:scenario3}
\end{table}

\begin{table}[!t]
\centering
\scalebox{0.9}{
\begin{tabular}{l|ccc}
\toprule
M & Untargeted ACC & Targeted ASR & Targeted ACC \\ \midrule
100 & 74.37$\pm$3.21 & 11.85$\pm$1.74 & 75.48$\pm$3.10 \\
200 & 73.74$\pm$3.27 & 12.18$\pm$1.57 & 74.23$\pm$2.43 \\
500 & 74.82$\pm$3.08 & 11.76$\pm$1.82 & 74.32$\pm$2.64 \\
1000 & 74.66$\pm$2.63 & 11.63$\pm$1.85 & 74.26$\pm$2.77 \\
\bottomrule
\end{tabular}}
\caption{Results varying the number of synthetic samples ($M$)\looseness=-1}
\label{tab:appendix_test_batch_size}
\end{table}


\begin{figure*}[!ht]
    \centering
    \includegraphics[height=7.4cm,width=0.9\linewidth]{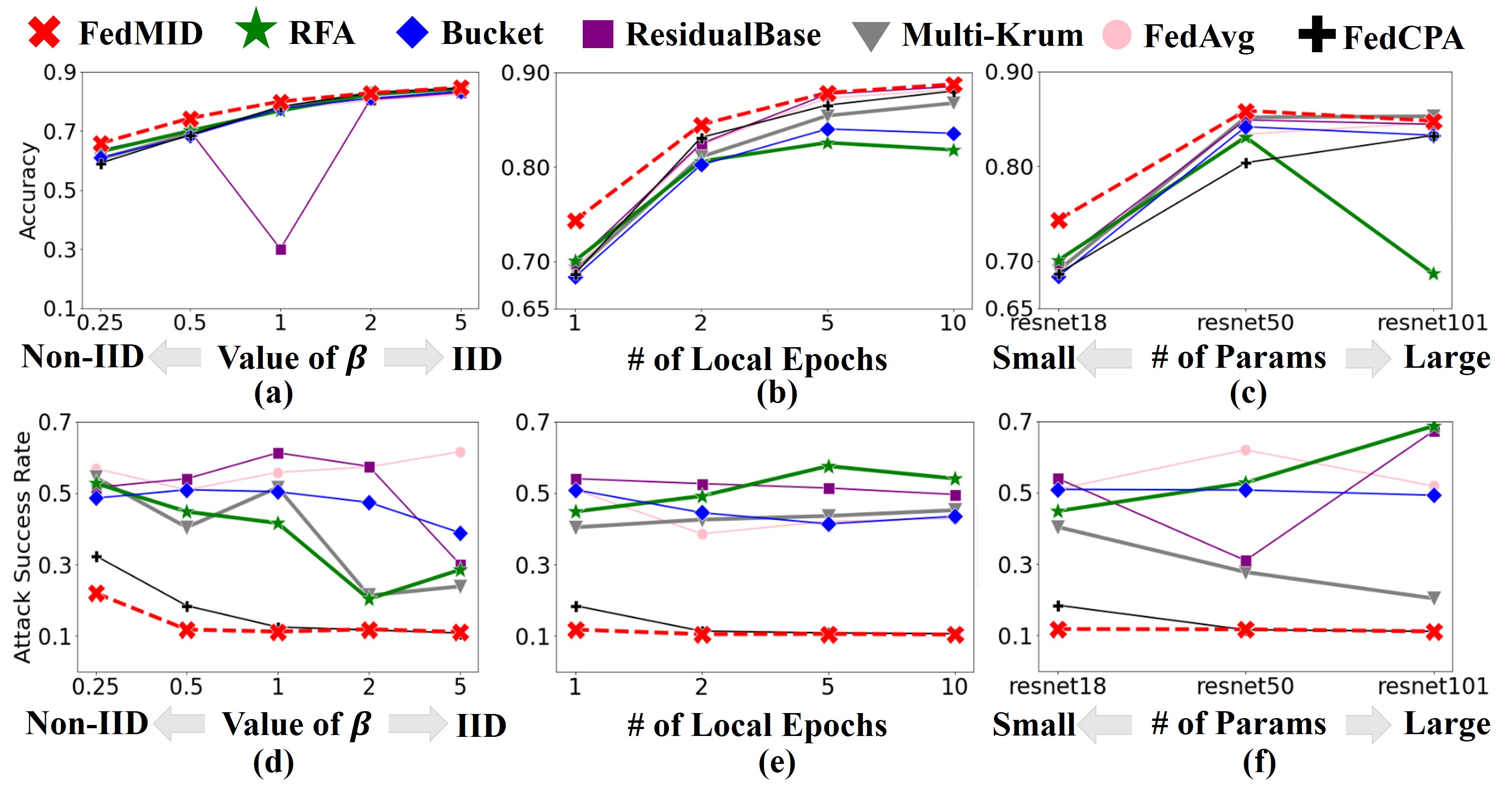}
    \caption{Robustness analysis under targeted attack scenarios with varying simulation parameters: (a,d): Effect of non-iidness, (b,e): Effect of number of local epochs, and (c,f): Effect of backbones. ACC and ASR among defense methods is shown.  \looseness=-1}
    \vspace*{-2mm}
    \label{fig:robust_acc_asr}    
\end{figure*}

\subsection{Robustness Test}
To evaluate the robustness, we conduct experiments under various simulation settings, such as the level of non-IIDness ($\beta$), the number of local epochs, varying backbone networks, and the number of synthetic samples. We use CIFAR-10 for all experiments and the same targeted attack settings from \textit{Scenario 1}.
Fig.~\ref{fig:robust_acc_asr} depicts the results comparing our method with the leading four baselines and the case with no defense (i.e., FedAvg). \model{} consistently demonstrates superior performance to all baselines, whereas parameter-based approaches show unstable performance over some parameter settings.
\suw{For example, ResidualBase and RFA showed abrupt accuracy fluctuations in some settings, which are depicted as high standard deviations in Appendix Tables~\ref{tab:appendix_iid_tar_acc} and~\ref{tab:appendix_model_size_taracc}.} 
Table~\ref{tab:appendix_test_batch_size} presents the effect of varying the number of synthetic samples ($M$) to produce intermediate outputs. Our proposed method also maintains stable performance across $M$ values, \suw{showing that \model{} still performs well even in the absence of a large synthetic dataset (i.e. less than 100).} \looseness=-1

\vspace{-1mm}
\section{Related Works}
\paragraph{Federated Learning}
Since the proposal of Federated Learning by FedAvg~\cite{mcmahan2017communication}, numerous studies have been conducted in this field. 
Several lines of research have concentrated on addressing heterogeneous settings~\cite{li2020fedprox,horvth2021fjord,karimireddy2020scaffold,ye2023feddisco}, reducing the number of communication rounds~\cite{yurochkin2019bayesian,lee2023layer}, and enhancing memory efficiency~\cite{bacon2016,caldas2018expanding,hamer20fedboost}. These endeavors strive to navigate the challenges presented by distributed settings. \looseness=-1

\cutparagraphup
\paragraph{Attack on Federated Learning}
Poisoning attacks~\cite{baruch2019little} constitute a major threat in federated learning, where attackers aim to either degrade the overall model performance (untargeted attack)~\cite{steinhardt2017certified}, or manipulate the model to behave in a specific desired manner (targeted or backdoor attacks)~\cite{shafahi2018poison,sun2019can,bagdasaryan2020backdoor,xie2020dba}.
More sophisticated attacks have been designed to evade known defense strategies.
For instance, in these advanced attacks, adversaries may be distributed across the network~\cite{xie2020dba}, or attackers are aware of benign clients' updates and exploit this information to introduce malicious updates~\cite{baruch2019little,fang2020local,shejwalkar2021manipulating}. \looseness=-1

\cutparagraphup
\paragraph{Defense Strategies}
The robustness of federated learning algorithms against poisoning attacks remains an open research question.
Numerous studies have proposed parameter-based defense methods to counter such attacks in federated learning. Simple yet effective defense methods employ coordinate-wise aggregation rules such as median~\cite{xie2018generalized}, and trimmed mean~\cite{yin2018byzantine}, which offer statistical robustness.
Similarly, the Residual Base method~\cite{fu2019attack} inspects the confidence of coordinate values in model parameters. 
The RFA method~\cite{pillutla2022robust} utilizes the geometric median of client weights for robust updating.
A recent study~\cite{karimireddy2022byzantinerobust} introduced a sampling-based defense method that enables existing aggregation rules to function effectively in non-IID settings.

Some models devise measures of normality for local updates to filter out malicious updates. For instance, the Multi-Krum~\cite{blanchard2017machine}, FoolsGold~\cite{sun2019can}, FLTrust~\cite{cao2021fltrust}, DnC~\cite{shejwalkar2021manipulating}, and FedCPA~\cite{han2023towards} algorithms detect malicious clients by comparing local update vectors. 
Our work shares the objective of identifying malicious updates at the client level.  \looseness=-1

\section{Conclusion}
This paper presented a new paradigm for robust aggregation for federated learning to protect against poisoning attacks. In contrast to prior defense mechanisms that rely on local updates to model parameters to identify adversaries, our method directly compares the functional mappings among local models by examining the intermediate outputs derived from given inputs. 
We generate intermediate outputs using random synthetic datasets based on the standard normal distribution. This design decision mitigates privacy concerns, which is important for real-world applications.
Extensive experiments demonstrate that our proposed method demonstrates robust detection performance under a variety of attack scenarios and is resilient to varying simulation parameters.\looseness=-1
\bibliographystyle{ijcai24}
\bibliography{ijcai24}

\newpage
\appendix
\onecolumn

\section{Appendix A - Setting Details}
We present details of our experimental setting, attack scenarios, and baseline defense methods.

\subsection{Federated system details}
Following upon prior research in federated learning~\cite{li2021model,han2022fedx}, we employ ResNet-18 as our default backbone network. 
By default, we set the total number of clients and communication rounds to 20 and 100, respectively. 
To simulate a more realistic scenario, half of the clients (i.e., $N/2=10$) are randomly selected at the initial stage of each round.
The data distribution for class $c$ within each client is determined by the probability $p^c_i$, which is sampled from the Dirichlet distribution $Dir(N,\beta)$.
Here, the parameter $\beta$ controls the level of non-IIDness, where a lower $\beta$ value indicates a higher level of non-IIDness. As a default, we set $\beta$ to 0.5.
Each round of communication corresponds to one epoch of training for the local model. 
We employ the SGD optimizer with a learning rate of 0.01, a momentum of 0.9, and a weight decay parameter of 1e-5. The batch size is fixed at 64.
Additionally, we leverage data augmentation techniques such as random cropping, horizontal flipping, and color jittering to enhance model performance.

\subsection{Attack details}
We consider three attack scenarios: (1) attackers have no information about benign clients, (2) attackers have partial information about benign clients, and (3) attackers are aware of the defense methods and attempt to bypass them. For all scenarios, we implement both targeted and untargeted versions of attacks for evaluation. 
We describe details for each attack scenario below. \medskip

\begin{itemize}[nosep,leftmargin=1em,labelwidth=*,align=left]
    \item In \textit{Scenario 1}, we employ the label flipping attack for poisoning, as it can be executed without any information about benign clients. In untargeted attacks, attackers randomly flip the fraction of labels in their datasets to send noisy local updates. 
    In targeted attacks, attackers insert the same small markers in the fraction of the dataset and flip their labels to the single target class. The pollution ratio ($\gamma_p$) determines the fraction of corrupted input-target pairs in the attacker's dataset.
    We generate a backdoor noise input pattern, following the method outlined in existing literature~\cite{gu2017badnets}.
    The backdoor pattern measures 5$\times$5 and is positioned in the bottom-right corner of the input images. In the targeted setting, we utilize pollution ratios of 0.5 and 0.8, while pollution ratios of 0.8 and 1.0 are adopted in the untargeted setting. \medskip
    
    \item In \textit{Scenario 2}, we incorporate advanced attack strategies from existing literature, specifically the LIE strategy~\cite{baruch2019little}, applicable to both targeted and untargeted label flipping attacks. The LIE strategy allows attackers to utilize the mean ($\theta^t_b$) and the standard deviation ($\sigma^t_b$) of update parameters from benign clients to generate malicious updates ($\Delta^t_a$) defined as $\Delta^t_a = \theta^t_b + z\sigma^t_b$, where $z$ denotes the perturbation scaling factor. \looseness=-1 \medskip
    
    \item In \textit{Scenario 3}, attackers attempt to regularize intermediate outputs from their models to resemble those from the global model.
    The intuition behind this attack strategy is that the global model can be considered representative of all clients, and thus mimicking the global model can be helpful to circumvent the defense.
    Specifically, attackers train their models with the objective $L=L_{\text{adv}} + L_{\text{reg}}$, where $L_{\text{adv}}$ is the objective for malicious attacks, either untargeted or targeted, as the one used in Scenario 1, and $L_{reg}$ is a regularization term defined as follows:
    \begin{equation}
    L_{\text{reg}}= {1 \over |\mathcal{D}_G|} \sum_{\mathbf{x}_k \in \mathcal{D}_G}||\mathbf{z}^{(l)}_{k,b}-\mathbf{z}^{(l)}_{k,a}||_2.
    \end{equation}
    In this definition, $\mathbf{z}^{(l)}_{k,b}$ is the $l$-th intermediate output of input $\mathbf{x}_k$ from the global model, and $\mathbf{z}^{(l)}_{k,a}$ is from the malicious model.
    For simplicity, we do not introduce the adjusting factor between two objectives, $L_{\text{adv}}$ and $L_{\text{reg}}$. 
    \looseness=-1
\end{itemize}

\subsection{Overall procedure of \model{}}
We below describe the overall procedure of \model{} in Algorithm~\ref{alg:mainalg}.
\begin{algorithm}[h!]
\DontPrintSemicolon
\SetAlgoLined
\SetNoFillComment 
\caption{Aggregation process of \model{}}\label{alg:mainalg}
\begin{flushleft}
\small{
\textbf{Require:} Number of clients $N$, a set of layers $\mathcal{L}$, a size of the synthetic dataset $M$, global model parameters $\phi^t$, and local update vectors $\Delta_i^t$ for each participating client $i\in [1..N]$ in the round.  \medskip

$\mathcal{D}_G \sim \mathcal{N}(0,1)$, where $|D_G|=M$  \Comment{Generate synthetic dataset} \\

\For{each client $i \in [1..N]$}
{
    $\theta_i^t = \phi^t + \Delta_i^t$  \\
    \For {each sample $\mathbf{x}_k \in \mathcal{D}_G$}
    {
        $\mathbf{z}_k \leftarrow f_{\theta_i^t}(\mathbf{x})$ \Comment{Get intermediate outputs}
    }
    \For {$k_1 \in [1..M]$ and $k_2 \in [1..M]$ and $l \in \mathcal{L}$}
    {
    $S^{(l)}_i [k_1, k_2] = || \mathbf{z}^{(l)}_{k_1} - \mathbf{z}^{(l)}_{k_2} ||_2,$ \Comment{Compute distance matrix}
    }
}
\For{each client $i \in [1..N]$}
{
    \For{each layer $l \in \mathcal{L}$}
    {
        \For{each client $j \neq i, j \in [1..N]$ } 
        { 
        $\text{Dist}^{(l)}(i, j) = {1 \over M^2} \sum_{k_1=1}^M \sum_{k_2=1}^M  |S^{(l)}_i [k_1, k_2] - S^{(l)}_j [k_1, k_2]|$  \Comment{Measure difference}
        }
        $\text{Anom}^{(l)}(i) = \text{Median}(\{ \text{Dist}^{(l)}(i, j) | j \in \mathcal{C}^t \})$ 
    }
    $\mathcal{N}(i) = - {1 \over |\mathcal{L}|} \sum_{l \in \mathcal{L}} \text{Normalize}(\text{Anom}^{(l)}(i))$ \Comment{Assessing normality score} \\
    $\mathcal{\Tilde{N}}(i) =\text{Scale}(\mathcal{N}(i))$ \\
    $a_i = \text{Clamp}_{0\sim 1}(\ln {\mathcal{\Tilde{N}}(i) \over {1 - \mathcal{\Tilde{N}}(i)}} + 0.5)$ \\
    $\mathcal{A}{(i)}={{a_i}\over \sum_{j \in \mathcal{C}^t}\mathbf{1}(a_j >0)}$
}
$\phi^{t+1} = \phi^t + \sum_{i\in [1..N]} \mathcal{A}(i) \cdot \Delta_i^t$ \Comment{Attack-tolerant aggregation}
}
\end{flushleft}
\end{algorithm}

\subsection{Baselines}
We describe each baseline defense strategy utilized in our evaluation.
We refer to the original works' settings and details to implement all baselines. \looseness=-1 \medskip

\begin{itemize}[nosep,leftmargin=1em,labelwidth=*,align=left]
    \item \textbf{FedAvg}: This is the standard federated learning algorithm that does not incorporate any specific methods to defend against attacks. It averages local updates from all participating clients during a round to produce the global model. \looseness=-1 \smallskip
   
  \item \textbf{Median}: Instead of the conventional averaging operation for aggregating local update vectors, this method employs a coordinate-wise median operation. \smallskip
   
  \item \textbf{Trimmed mean}: Similar to the Median strategy, this approach uses a coordinate-wise trimmed mean operation in place of the averaging operation while aggregating local update vectors. Both the Median and Trimmed Mean methods are capable of mitigating the influence of extreme values within each coordinate of local updates.\smallskip
   
  \item \textbf{Multi-Krum}: This strategy, with the assumption that the central server is already aware of the maximum number of attackers in the system, iteratively removes the local update vector farthest away in terms of Euclidean distance. \smallskip 
   
  \item \textbf{Foolsgold}: This method aims to identify coordinated attacks by filtering out groups of local update vectors that show abnormally high similarity. \smallskip 
   
  \item \textbf{ResidualBase}: This strategy executes residual analysis and computes the confidence of each local update. Updates exhibiting low confidence are discarded. This method's confidence interval and clipping threshold are set to 2.0 and 0.05, respectively. \smallskip
   
  \item \textbf{RFA}: This strategy proposes using a geometric median operation for robust aggregation that tolerates attacks. The smoothing parameter for RFA is set to 1e-6, and the maximum number of Weiszfeld iterations is set to 100. \smallskip
  
  \item \textbf{DnC}: This strategy uses a singular vector decomposition (SVD) based spectral method for detecting outliers among local update vectors. DnC iteratively performs random sampling for local updates and computes outlier scores. The number of iterations is set to 1, filtering fraction $c$ is set to 1, the dimension of subsamples is set to 10,000, and the number of malicious clients is set to the number of expected malicious clients ($N \times$ attacker ratio). \smallskip
  
  \item \textbf{Bucket}: This method proposes the utilization of a bucketing step operation, which can be applied in conjunction with any existing aggregation rules. The hyperparameter $s$ for bucketing is set to 2, and RFA is employed for aggregation rules.  \smallskip 
  
  \item \textbf{FLTrust}: This method necessitates the use of a root dataset, sampled from the same distribution as the entire training dataset, to enable robust aggregation. Specifically, for each communication round, the global model is trained to obtain a benign update gradient ($\mathbf{g}_0$). Updates from each local client ($\mathbf{g}_i$) are then normalized based on the similarity between global and local updates, denoted as $sim(\mathbf{g}_0,\mathbf{g}_i)$. This similarity-based update mechanism assigns lower weights to malicious updates and higher weights to benign updates, thereby enhancing the model's resilience to adversarial influences. In our implementation, we utilized 100 samples from the training dataset as the root dataset, and we set the number of global epochs to be equal to the number of local epochs.  
  \looseness=-1 \smallskip

  \item \textbf{FedCPA}: \suw{This method measures the significance of model parameters ($p_i$) by computing the product of the update magnitude ($\Delta_i=\theta_i-\phi_i$) and the weight magnitude ($p_i=\left\vert \Delta_i\cdot\theta_i\right\vert$). Utilizing this parameter significance, the approach selects the top-k and bottom-k significant parameters for comparison using Jaccard similarity. Subsequently, it evaluates the Spearman correlation between the importance values. By integrating the Jaccard similarity and Spearman correlation, the method computes the similarity between each client's update. This similarity measure is then normalized to facilitate the aggregation process.}

\end{itemize}

\newpage

\section{Appendix B - Full Results}
We report the entire experimental results for our performance evaluation and robustness test over four benchmark datasets.
We present the top-1 classification accuracy of the global model over the test dataset.
To ensure a stable evaluation, we calculate the average and standard deviation of accuracy (ACC) across the last 10 epochs.
In the case of targeted attacks, we also report the attack success rate (ASR) for the last 10 epochs.

\subsection{Experimental results under Scenario 1}
\begin{table*}[!ht]
\caption{Experimental results under untargeted label flipping attacks. ACC is reported.}
\centering
\begin{tabular}{l|cccc}
\hline
\multicolumn{1}{c|}{\begin{tabular}[c]{@{}c@{}} Untargeted (ACC)\end{tabular}} & CIFAR10 & CIFAR100 & SVHN & \begin{tabular}[c]{@{}c@{}}TinyImageNet\end{tabular} \\ \hline
FedAvg   & 67.82$\pm $ 3.49          & 42.56$\pm $   1.60          & 90.64$\pm $  1.80          & 33.01$\pm $ 4.76          \\
Median       & 59.77$\pm $ 3.16          & 36.59$\pm $ 1.09          & 89.91$\pm $ 1.55          & 28.66$\pm $ 4.73          \\
Trimmed Mean & 71.43$\pm $ 3.47          & 43.87$\pm $ 1.36          & 91.00$\pm $ 1.49          & 34.11$\pm $ 3.73          \\
Multi-Krum   & 72.75$\pm $ 3.61          &  {47.28$\pm $ 0.67} & 92.64$\pm $ 0.99          & 35.88$\pm $ 2.22          \\
FoolsGold    & 18.56$\pm $ 7.53          & 12.75$\pm $ 8.04          & 47.62$\pm $ 19.76         & 4.64$\pm $ 3.36           \\
ResidualBase & 73.61$\pm $ 3.40          & 44.73$\pm $ 1.25          & 92.08$\pm $ 1.03          &  {36.00$\pm $ 3.38} \\
RFA          & 72.65$\pm $ 2.31          & 37.89$\pm $ 0.46          & 92.65$\pm $ 0.96          & 36.46$\pm $ 0.78          \\
DnC          &    72.35$\pm$4.80	 & 47.85$\pm$0.67	& 90.68$\pm$3.14	& 35.89$\pm$3.48          \\
Bucket          &  70.12$\pm$2.76	 & 48.45$\pm$0.18	& 90.77$\pm$1.38	& 37.89$\pm$0.46     \\
FLTrust       &  65.61$\pm$2.44& 39.74$\pm$ 2.46          &  88.42$\pm $ 1.97 & 33.13$\pm $ 2.38          \\
FedCPA       &  74.86$\pm$3.30& 42.87$\pm$ 0.98          &  93.17$\pm $ 0.72 & 36.84$\pm $ 1.53          \\
FedMID       &  {74.82$\pm $ 3.08} & 48.08$\pm $ 0.88          &  {93.04$\pm $ 0.43} & 36.71$\pm $ 0.90          \\ \hline
\end{tabular}
\label{tab:appendix_main_unt_full}
\end{table*}
\begin{table*}[!ht]
\caption{Experimental results under targeted label flipping attacks. ASR is reported.}
\centering
\begin{tabular}{l|cccc}
\hline
\multicolumn{1}{c|}{\begin{tabular}[c]{@{}c@{}}Targeted (ASR)\end{tabular}} &
  CIFAR10 &
  CIFAR100 &
  SVHN &
  \begin{tabular}[c]{@{}c@{}} TinyImageNet \end{tabular} \\ \hline
FedAvg   & 50.90$\pm $25.09 & 48.08$\pm $48.09 & 21.97$\pm $2.21  & 96.12$\pm $1.34  \\
Median       & 70.62$\pm $16.51 & 8.18$\pm $8.81   & 23.58$\pm $3.39  & 96.15$\pm $0.59  \\
Trimmed Mean & 18.99$\pm $10.29 & 36.51$\pm $37.62 & 21.42$\pm $1.78  & 97.00$\pm $0.81  \\
Multi-Krum   & 40.39$\pm $21.85 & 0.56$\pm $0.36   & 23.37$\pm $4.06  & 18.97$\pm $13.91 \\
FoolsGold    & 46.84$\pm $34.83 & 70.91$\pm $42.00 & 32.30$\pm $27.72 & 69.06$\pm $43.20 \\
ResidualBase & 54.00$\pm $27.50 & 21.57$\pm $26.67 & 21.86$\pm $2.34  & 96.18$\pm $0.81  \\
RFA          & 44.79$\pm $21.58 & 0.72$\pm $0.37   & 22.04$\pm $2.01  & 11.39$\pm $5.80  \\
DnC          &  36.03$\pm$16.60	& 1.44$\pm$1.85	& 21.52$\pm$1.84	& 32.57$\pm$8.99              \\
Bucket          & 50.91$\pm$21.97	& 16.71$\pm$14.24	& 20.97$\pm$1.09	& 93.20$\pm$2.53 \\
FLTrust &
  {43.42$\pm $16.41} &
  {44.07$\pm $36.21} &
  {21.35$\pm $0.87} &
  {96.31$\pm $0.82} \\
FedCPA &
  {18.45$\pm $5.19} &
  {0.55$\pm $0.28} &
  {21.00$\pm $1.17} &
  {0.65$\pm $0.15} \\
FedMID &
  {11.76$\pm $1.82} &
  {0.62$\pm $0.33} &
  {20.93$\pm $1.21} &
  {0.62$\pm $0.29} \\ \hline
\end{tabular}
\label{tab:appendix_targeted_asr}
\end{table*}
\begin{table*}[!ht]
\caption{Experimental results under targeted label flipping attacks. ACC is reported.}
\centering
\begin{tabular}{l|cccc}
\hline
\multicolumn{1}{c|}{Targeted (ACC)} & CIFAR10 & CIFAR100 & SVHN & TinyImageNet \\ \hline
FedAvg   & 69.31$\pm $3.74          & 31.12$\pm $11.64         &  {92.52$\pm $0.93} &  {38.79$\pm $1.12} \\
Median       & 62.37$\pm $3.32          & 35.74$\pm $1.57          & 89.99$\pm $1.53          & 31.49$\pm $0.99          \\
Trimmed Mean & 71.42$\pm $2.77          & 35.75$\pm $8.36          & 91.67$\pm $1.25          & 37.94$\pm $1.12          \\
Multi-Krum   & 68.97$\pm $2.21          & 47.09$\pm $0.72          & 90.70$\pm $2.33          & 36.27$\pm $1.78          \\
FoolsGold    & 49.02$\pm $9.46          & 9.23$\pm $1.07           & 69.84$\pm $24.56         & 28.51$\pm $4.27          \\
ResidualBase & 69.85$\pm $3.59          & 39.45$\pm $4.62          & 92.45$\pm $0.81          & 38.61$\pm $0.47          \\
RFA          & 70.06$\pm $3.37          &  {47.44$\pm $0.67} & 91.79$\pm $1.44          & 36.25$\pm $1.05          \\
DnC          &   70.03$\pm$3.00	& 46.76$\pm$0.87	& 91.57$\pm$0.90	& 37.15$\pm$0.83        \\
Bucket          &  68.34$\pm$4.21	& 42.55$\pm$0.90	& 92.15$\pm$0.75	& 37.78$\pm$1.11   \\
FLTrust       &  {69.91$\pm $2.11} & 35.94$\pm $5.03          & 91.69$\pm $0.48          & 38.64$\pm $0.36          \\ 
FedCPA       &  {68.68$\pm $4.18} & 39.80$\pm $1.27          & 93.05$\pm $0.38          & 38.11$\pm $0.73          \\ 
FedMID       &  {74.32$\pm $2.64} & 48.17$\pm $0.36          & 93.05$\pm $0.38          & 36.07$\pm $1.04          \\ \hline
\end{tabular}
\label{tab:appendix_tar_acc}
\end{table*}

\newpage
\subsection{Experimental results under Scenario 2}
\begin{table*}[!ht]
\caption{Experimental results under LIE attacks. UNT and TAR refer to untargeted and targeted settings, respectively. ACC for untargeted attacks and ACC/ASR for targeted attacks are reported.}
\centering
\begin{tabular}{l|l|ll}
\hline
\multicolumn{1}{c|}{\multirow{2}{*}{LIE}} & \multicolumn{1}{c|}{UNT} & \multicolumn{2}{c}{TAR}                           \\ \cline{2-4} 
\multicolumn{1}{c|}{}                     & \multicolumn{1}{c|}{ACC} & \multicolumn{1}{c}{ACC} & \multicolumn{1}{c}{ASR} \\ \hline
FedAvg   & 71.63$\pm$2.83 & 69.80$\pm$3.35 & 51.95$\pm$23.42 \\
Median       & 66.75$\pm$4.55 & 62.40$\pm$4.29 & 69.98$\pm$11.96 \\
Trimmed Mean & 69.91$\pm$3.26 & 69.25$\pm$3.85 & 55.90$\pm$24.71 \\
Multi-Krum   & 66.90$\pm$3.64 & 70.20$\pm$2.94 & 23.17$\pm$14.03 \\
FoolsGold    & 70.48$\pm$2.03 & 62.24$\pm$4.07 & 40.22$\pm$13.35 \\
ResidualBase & 71.01$\pm$2.82 & 69.09$\pm$3.53 & 53.40$\pm$27.61 \\
RFA          & 59.43$\pm$9.73 & 70.19$\pm$3.06 & 46.02$\pm$22.02 \\
DnC          &  54.52$\pm$3.51 &	69.19$\pm$3.47	& 42.59$\pm$21.82                 \\
Bucket          &   70.59$\pm$2.41	& 70.17$\pm$4.06	& 49.52$\pm$21.60             \\
FLTrust                                    &  {72.34$\pm$2.05}     &  {69.94$\pm$1.26}    &  {52.83$\pm$18.58}    \\
FedCPA          & 63.95$\pm$7.65 & 60.60$\pm$3.14 & 23.59$\pm$18.64 \\
FedMID                                    &  {71.42$\pm$2.57}     &  {72.83$\pm$2.23}    &  {11.88$\pm$1.64}    \\ 
\hline
\end{tabular}
\label{tab:appendix_LIE}
\end{table*}

\newpage 
\subsection{Robustness test results}
To assess the robustness of our proposed method, we conduct experiments under various attack scenarios with different simulation parameters, including the level of non-IIDness ($\beta$), the number of local epochs, varying backbone networks, the attacker ratio, the number of clients, and under the different federated learning algorithm - FedProx. 
We use the CIFAR-10 dataset for all experiments and the same targeted attack settings as in Scenario 1. \looseness=-1 \medskip

\noindent
\textbf{Varying the level of non-IIDness}
\begin{table}[H]
\vspace{-2mm}
\caption{Results varying the level of non-IIDness under untargeted attacks. ACC is reported. \looseness=-1}
\centering
\begin{tabular}{l|lllll}
\hline
\multicolumn{1}{c|}{\multirow{2}{*}{Untargeted (ACC)}} &
  \multicolumn{5}{c}{Level of non-iidness($\beta$)} \\ \cline{2-6} 
\multicolumn{1}{c|}{} &
  \multicolumn{1}{c}{0.25} &
  \multicolumn{1}{c}{0.5} &
  \multicolumn{1}{c}{1} &
  \multicolumn{1}{c}{2} &
  \multicolumn{1}{c}{5} \\ \hline
FedAvg   & 62.29$\pm$4.16  & 67.82$\pm$3.49 & 73.52$\pm$3.23  & 74.18$\pm$5.27  & 75.95$\pm$5.71  \\
Median       & 52.19$\pm$4.21  & 59.77$\pm$3.16 & 74.56$\pm$2.31  & 79.34$\pm$2.05  & 79.85$\pm$2.25  \\
Trimmed Mean & 61.96$\pm$4.63  & 71.43$\pm$3.47 & 73.35$\pm$3.64  & 77.66$\pm$5.29  & 79.87$\pm$5.04  \\
Multi-Krum   & 63.03$\pm$4.47  & 72.75$\pm$3.61 &  {79.77$\pm$1.04}  &  {82.42$\pm$1.68}  &  {84.30$\pm$0.89}  \\
FoolsGold    & 29.44$\pm$11.26 & 18.56$\pm$7.53 & 30.53$\pm$13.35 & 29.89$\pm$12.74 & 23.05$\pm$11.89 \\
ResidualBase & 63.16$\pm$4.04  & 73.61$\pm$3.40 & 75.61$\pm$2.97  & 79.96$\pm$3.31  & 81.37$\pm$4.29  \\
RFA          & 62.79$\pm$3.46  & 72.65$\pm$2.31 & 77.87$\pm$2.09  & 80.93$\pm$1.47  & 84.10$\pm$1.12  \\
DnC          & 64.40$\pm$3.97	& 72.35$\pm$4.80	& 78.97$\pm$1.83	& 82.01$\pm$1.69	& 84.16$\pm$1.51  \\
Bucket          &  58.92$\pm$6.23	& 70.12$\pm$2.76	& 70.81$\pm$5.72	& 80.60$\pm$2.19	& 83.92$\pm$0.82  \\
FLTrust       &  {58.77$\pm$5.13}  &  {65.61$\pm$2.44} & 64.91$\pm$5.16  & 71.67$\pm$4.72  & 73.79$\pm$7.65  \\
FedCPA       &  {61.98$\pm$5.93}  &  {74.86$\pm$3.30} & 75.01$\pm$2.03  & 81.14$\pm$1.47  & 83.39$\pm$0.75  \\
FedMID       &  {66.65$\pm$3.83}  &  {74.82$\pm$3.08} & 80.33$\pm$0.66  & 82.34$\pm$1.76  & 84.28$\pm$0.83  \\ \hline
\end{tabular}
\label{tab:appendix_iid_unt}
\end{table}
\begin{table}[H]
\caption{Results varying the level of non-IIDness under targeted attacks. ASR is reported. \looseness=-1}
\centering
\begin{tabular}{l|lllll}
\hline
\multicolumn{1}{c|}{\multirow{2}{*}{Targeted (ASR)}} & \multicolumn{5}{c}{Level of non-iidness($\beta$)}                               \\ \cline{2-6} 
\multicolumn{1}{c|}{} & \multicolumn{1}{c}{0.25} & \multicolumn{1}{c}{0.5} & \multicolumn{1}{c}{1} & \multicolumn{1}{c}{2} & \multicolumn{1}{c}{5} \\ \hline
FedAvg                                         & 56.87$\pm$34.98 & 50.90$\pm$25.09 & 55.79$\pm$32.87 & 57.32$\pm$28.94 & 61.57$\pm$31.04 \\
Median                                             & 60.89$\pm$25.18 & 70.62$\pm$16.51 & 60.19$\pm$11.69 & 68.27$\pm$10.41 & 70.06$\pm$14.20 \\
Trimmed Mean                                       & 52.92$\pm$32.03 & 18.99$\pm$10.29 & 56.72$\pm$29.70 & 56.72$\pm$29.70 & 23.86$\pm$15.19 \\
Multi-Krum                                         & 54.38$\pm$28.36 & 40.39$\pm$21.85 & 51.57$\pm$20.04 & 21.31$\pm$13.96 & 23.86$\pm$15.19 \\
FoolsGold                                          & 38.25$\pm$16.37 & 46.84$\pm$34.83 & 42.04$\pm$30.11 & 54.95$\pm$35.55 & 73.80$\pm$27.84 \\
ResidualBase                                       & 51.57$\pm$31.26 & 54.00$\pm$27.50 & 61.28$\pm$29.15 & 57.47$\pm$30.73 & 30.06$\pm$32.87 \\
RFA                                                & 52.72$\pm$24.73 & 44.79$\pm$21.58 & 41.58$\pm$12.39 & 20.30$\pm$4.17  & 28.49$\pm$3.73  \\
DnC          &     44.07$\pm$25.65	& 36.03$\pm$16.60	& 31.64$\pm$21.27	& 25.13$\pm$13.47	& 25.44$\pm$15.84\\
Bucket          &  48.57$\pm$34.17	& 50.91$\pm$21.97	& 50.37$\pm$26.30	&47.39$\pm$16.19	& 38.90$\pm$1.03 \\
FLTrust       &  {32.15$\pm$13.61}  &  {{43.42$\pm $16.41}} & 48.21$\pm$20.62  & 60.84$\pm$25.92  & 67.08$\pm$1.04  \\
FedCPA       &  {32.42$\pm$16.88}  &  {18.45$\pm$5.19} & 12.45$\pm$0.77  & 11.66$\pm$0.78  & 10.79$\pm$0.97  \\
FedMID                                             & 21.98$\pm$6.34  & 11.76$\pm$1.82  & 11.18$\pm$0.57  & 11.84$\pm$1.04  & 11.07$\pm$0.76  \\ \hline
\end{tabular}
\label{tab:appendix_iid_tar_asr}
\end{table}
\begin{table}[H]
\caption{Results varying the level of non-IIDness under targeted attacks. ACC is reported. \looseness=-1}
\centering
\begin{tabular}{l|lllll}
\hline
\multicolumn{1}{c|}{\multirow{2}{*}{Targeted (ACC)}} &
  \multicolumn{5}{c}{Level of non-iidness($\beta$)} \\ \cline{2-6} 
\multicolumn{1}{c|}{} &
  \multicolumn{1}{c}{0.25} &
  \multicolumn{1}{c}{0.5} &
  \multicolumn{1}{c}{1} &
  \multicolumn{1}{c}{2} &
  \multicolumn{1}{c}{5} \\ \hline
FedAvg   & 63.15$\pm$2.22 & 69.31$\pm$3.74 & 77.12$\pm$2.26  & 80.30$\pm$1.61  & 82.27$\pm$1.56 \\
Median       & 54.68$\pm$2.38 & 62.37$\pm$3.32 & 70.06$\pm$14.20 & 78.60$\pm$1.99  & 80.89$\pm$1.45 \\
Trimmed Mean & 63.24$\pm$2.38 & 71.42$\pm$2.77 & 61.95$\pm$31.21 & 80.33$\pm$1.72  & 82.35$\pm$1.18 \\
Multi-Krum   & 60.57$\pm$4.02 & 68.97$\pm$2.21 & 77.51$\pm$1.33  & 82.85$\pm$1.06  & 84.13$\pm$0.60 \\
FoolsGold    & 47.60$\pm$4.34 & 49.02$\pm$9.46 & 52.46$\pm$17.14 & 51.40$\pm$11.36 & 54.86$\pm$6.47 \\
ResidualBase & 63.56$\pm$1.91 & 69.85$\pm$3.59 & 30.06$\pm$32.87 & 80.63$\pm$1.21  & 82.93$\pm$0.74 \\
RFA          & 62.94$\pm$3.47 & 70.06$\pm$3.37 & 76.83$\pm$1.42  & 82.32$\pm$1.45  & 84.25$\pm$1.10 \\
DnC          & 61.88$\pm$3.40	& 70.03$\pm$3.00	& 78.30$\pm$1.36	& 82.37$\pm$1.50	& 84.46$\pm$1.55\\
Bucket          & 60.85$\pm$3.39 &	68.34$\pm$4.21 &	77.56$\pm$1.21 & 81.02$\pm$1.10	& 83.32$\pm$1.03 \\
FLTrust       &  {65.33$\pm$2.85}  &  {69.91$\pm $2.11} & 76.90$\pm$0.83  & 81.87$\pm$0.78  & 82.86$\pm$1.84  \\
FedCPA       &  {59.07$\pm$3.53}  &  {68.68$\pm$4.18} & 78.24$\pm$1.46  & 82.73$\pm$1.50  & 84.45$\pm$0.97  \\
FedMID       & 65.79$\pm$5.59 & 74.32$\pm$2.64 & 79.94$\pm$1.46  & 82.79$\pm$1.32  & 84.67$\pm$0.95 \\ \hline
\end{tabular}
\label{tab:appendix_iid_tar_acc}
\end{table}
\newpage 
\noindent
\textbf{Varying the number of local epochs}
\begin{table}[!h]
\caption{Result varying the number of local epochs in untargeted attacks. ACC is reported.}
\centering
\begin{tabular}{l|llll}
\hline
\multicolumn{1}{c|}{\multirow{2}{*}{Untargeted (ACC)}} & \multicolumn{4}{c}{\# of local epoch}                                                          \\ \cline{2-5} 
\multicolumn{1}{c|}{}                              & \multicolumn{1}{c}{1} & \multicolumn{1}{c}{2} & \multicolumn{1}{c}{5} & \multicolumn{1}{c}{10} \\ \hline
FedAvg   & 67.82$\pm$3.49 & 79.03$\pm$3.53  & 82.12$\pm$3.77  & 85.87$\pm$1.63  \\
Median       & 59.77$\pm$3.16 & 74.58$\pm$3.57  & 80.78$\pm$2.82  & 82.72$\pm$2.66  \\
Trimmed Mean & 71.43$\pm$3.47 & 79.70$\pm$2.83  & 84.25$\pm$2.79  & 82.65$\pm$4.25  \\
Multi-Krum   & 72.75$\pm$3.61 & 82.86$\pm$1.82  & 86.12$\pm$2.20  & 84.59$\pm$2.18  \\
FoolsGold    & 18.56$\pm$7.53 & 31.47$\pm$16.95 & 48.09$\pm$18.32 & 44.92$\pm$13.66 \\
ResidualBase & 73.61$\pm$3.40 & 81.33$\pm$1.99  & 85.74$\pm$1.88  & 87.85$\pm$0.86  \\
RFA          & 72.65$\pm$2.31 & 81.32$\pm$1.71  & 82.57$\pm$2.18  & 82.63$\pm$2.94  \\
DnC          & 72.35$\pm$4.80	& 82.93$\pm$2.26	& 86.67$\pm$2.11	& 87.69$\pm$0.80  \\
Bucket          &  70.12$\pm$2.76 &	80.22$\pm$2.33 &	82.68$\pm$2.18	& 85.52$\pm$0.73 \\
FLTrust       & 65.61$\pm$2.44 & 76.25$\pm$2.86  & 82.16$\pm$3.23  & 85.89$\pm$1.67  \\
FedCPA       &  {74.86$\pm$3.30} & 84.74$\pm$1.32  & 87.50$\pm$0.74  & 88.03$\pm$0.79  \\
FedMID       & 74.32$\pm$3.08 & 84.29$\pm$1.36  & 87.97$\pm$0.74  & 88.06$\pm$0.84  \\ \hline
\end{tabular}
\label{tab:appendix_local_epoch_unt}
\end{table}
%
\begin{table}[!h]
\caption{Result varying the number of local epochs in targeted attacks. ASR is reported.}
\centering
\begin{tabular}{l|llll}
\hline
\multicolumn{1}{c|}{\multirow{2}{*}{Targeted (ASR)}} & \multicolumn{4}{c}{\# of local epoch}                                                          \\ \cline{2-5} 
\multicolumn{1}{c|}{}                              & \multicolumn{1}{c}{1} & \multicolumn{1}{c}{2} & \multicolumn{1}{c}{5} & \multicolumn{1}{c}{10} \\ \hline
FedAvg   & 50.90$\pm$25.09 & 38.60$\pm$18.99 & 42.06$\pm$17.42 & 42.89$\pm$18.64 \\
Median       & 70.62$\pm$16.51 & 55.12$\pm$19.35 & 61.90$\pm$17.82 & 69.76$\pm$13.90 \\
Trimmed Mean & 18.99$\pm$10.29 & 49.56$\pm$22.43 & 45.72$\pm$22.51 & 52.91$\pm$24.09 \\
Multi-Krum   & 40.39$\pm$21.85 & 42.56$\pm$15.41 & 43.58$\pm$19.52 & 45.19$\pm$19.01 \\
FoolsGold    & 46.84$\pm$34.83 & 50.47$\pm$27.06 & 68.49$\pm$37.92 & 64.52$\pm$43.30 \\
ResidualBase & 54.00$\pm$27.50 & 52.64$\pm$24.95 & 51.38$\pm$23.57 & 49.60$\pm$24.58 \\
RFA          & 44.79$\pm$21.58 & 49.15$\pm$15.69 & 57.55$\pm$25.59 & 54.05$\pm$29.43 \\
DnC          & 36.03$\pm$16.60 &	30.08$\pm$10.64	& 25.61$\pm$17.94 &	29.41$\pm$9.23       \\
Bucket          &   50.91$\pm$21.97	& 44.49$\pm$17.65	& 41.38$\pm$22.10	& 43.46$\pm$23.89  \\
FLTrust       & 43.42$\pm$16.41 & 43.83$\pm$18.01  & 47.21$\pm$18.01  & 51.52$\pm$16.50  \\
FedCPA       &  {18.45$\pm$5.19} & 11.33$\pm$1.39  & 10.83$\pm$0.79  & 10.61$\pm$0.68  \\
FedMID       & 11.76$\pm$1.82  & 10.48$\pm$0.73  & 10.54$\pm$0.76  & 10.39$\pm$0.55  \\ \hline
\end{tabular}
\label{tab:appendix_targeted_asr_epoch}
\end{table}
%
\begin{table}[!h]
\caption{Result varying the number of local epochs in targeted attacks. ACC is reported.}
\centering
\begin{tabular}{l|llll}
\hline
\multicolumn{1}{c|}{\multirow{2}{*}{Targeted (ACC)}} & \multicolumn{4}{c}{\# of local epoch}                                                          \\ \cline{2-5} 
\multicolumn{1}{c|}{}                              & \multicolumn{1}{c}{1} & \multicolumn{1}{c}{2} & \multicolumn{1}{c}{5} & \multicolumn{1}{c}{10} \\ \hline
FedAvg   & 69.31$\pm$3.74 & 82.32$\pm$1.49 & 87.29$\pm$0.92  & 88.15$\pm$0.96  \\
Median       & 62.37$\pm$3.32 & 77.78$\pm$2.77 & 85.31$\pm$1.15  & 86.23$\pm$0.80  \\
Trimmed Mean & 71.42$\pm$2.77 & 82.53$\pm$1.39 & 87.40$\pm$0.81  & 88.76$\pm$0.59  \\
Multi-Krum   & 68.97$\pm$2.21 & 81.05$\pm$1.44 & 85.41$\pm$1.85  & 86.74$\pm$1.53  \\
FoolsGold    & 49.02$\pm$9.46 & 63.09$\pm$8.89 & 72.41$\pm$10.18 & 70.15$\pm$10.95 \\
ResidualBase & 69.85$\pm$3.59 & 82.45$\pm$1.21 & 87.68$\pm$0.70  & 88.50$\pm$0.71  \\
RFA          & 70.06$\pm$3.37 & 80.58$\pm$1.69 & 82.56$\pm$2.64  & 81.80$\pm$2.44  \\
DnC          &    70.03$\pm$3.00	& 82.28$\pm$2.97	& 87.38$\pm$0.79	& 88.35$\pm$0.93     \\
Bucket          & 68.34$\pm$4.21	& 80.20$\pm$2.86	& 84.00$\pm$3.19	& 83.54$\pm$4.24     \\
FLTrust       & 69.91$\pm$2.11 & 82.18$\pm$1.36  & 87.14$\pm$0.90  & 88.26$\pm$0.74  \\
FedCPA       & {68.68$\pm$4.18} & 83.11$\pm$1.29  & 85.63$\pm$0.78  & 88.01$\pm$0.83  \\
FedMID       & 74.32$\pm$2.64 & 84.47$\pm$1.34 & 87.83$\pm$0.86  & 88.74$\pm$0.65  \\ \hline
\end{tabular}
\label{tab:appendix_targeted_acc}
\end{table}
\newpage
\noindent
\textbf{Varying backbone networks}
\begin{table}[!ht]
\caption{Results varying the backbone network in untargeted attacks. ACC is reported.}
\centering
\begin{tabular}{l|lll}
\hline
\multicolumn{1}{c|}{\multirow{2}{*}{Untargeted (ACC)}} & \multicolumn{3}{c}{Model Size}                                                              \\ \cline{2-4} 
\multicolumn{1}{c|}{}                              & \multicolumn{1}{c}{Resnet18} & \multicolumn{1}{c}{Resnet50} & \multicolumn{1}{c}{Resnet101} \\ \hline
FedAvg   & 67.82$\pm$3.49 & 78.99$\pm$4.07  & 81.17$\pm$3.01  \\
Median       & 59.77$\pm$3.16 & 77.89$\pm$3.64  & 74.74$\pm$4.44  \\
Trimmed Mean & 71.43$\pm$3.47 & 81.04$\pm$3.20  & 81.01$\pm$2.19  \\
Multi-Krum   & 72.75$\pm$3.61 & 84.95$\pm$1.52  & 84.51$\pm$2.12  \\
FoolsGold    & 18.56$\pm$7.53 & 47.09$\pm$20.53 & 36.22$\pm$24.32 \\
ResidualBase & 73.61$\pm$3.40 & 82.69$\pm$1.68  & 82.76$\pm$2.04  \\
RFA          & 72.65$\pm$2.31 & 84.09$\pm$1.45  & 83.52$\pm$1.31  \\
DnC          &   72.35$\pm$4.80 & 83.69$\pm$2.03	& 82.83$\pm$4.09         \\
Bucket          &   70.12$\pm$2.76	& 82.22$\pm$3.47	& 83.26$\pm$2.19 \\
FLTrust          &   65.61$\pm$2.44	& 61.61$\pm$0.48	& 69.64$\pm$3.41 \\
FedCPA       &  {74.86$\pm$3.30} & 82.85$\pm$1.54  & 84.93$\pm$1.39 \\
FedMID       & 74.82$\pm$3.08 & 86.05$\pm$0.87  & 85.22$\pm$1.29  \\ \hline
\end{tabular}
\label{tab:appendix_model_size_unt}
\end{table}
%
\begin{table}[!ht]
\caption{Results varying the backbone network in targeted attacks. ASR is reported.}
\centering
\begin{tabular}{l|lll}
\hline
\multicolumn{1}{c|}{\multirow{2}{*}{Targeted (ASR)}} & \multicolumn{3}{c}{Model Size}                                                              \\ \cline{2-4} 
\multicolumn{1}{c|}{}                              & \multicolumn{1}{c}{Resnet18} & \multicolumn{1}{c}{Resnet50} & \multicolumn{1}{c}{Resnet101} \\ \hline
FedAvg   & 50.90$\pm$25.09 & 61.97$\pm$16.48 & 51.82$\pm$16.16 \\
Median       & 70.62$\pm$16.51 & 63.22$\pm$22.38 & 62.26$\pm$23.35 \\
Trimmed Mean & 18.99$\pm$10.29 & 22.32$\pm$9.22  & 73.00$\pm$20.61 \\
Multi-Krum   & 40.39$\pm$21.85 & 27.73$\pm$7.71  & 20.34$\pm$4.24  \\
FoolsGold    & 46.84$\pm$34.83 & 38.77$\pm$28.89 & 52.07$\pm$18.41 \\
ResidualBase & 54.00$\pm$27.50 & 30.97$\pm$14.37 & 67.26$\pm$18.48 \\
RFA          & 44.79$\pm$21.58 & 52.75$\pm$23.10 & 68.68$\pm$17.86 \\
DnC          &   36.03$\pm$16.60	& 51.65$\pm$11.97	& 19.58$\pm$20.14             \\
Bucket          &   50.91$\pm$21.97	& 50.72$\pm$13.32	& 49.24$\pm$10.82    \\
FLTrust          &   43.42$\pm$16.41	& 11.61$\pm$0.35	& 10.98$\pm$0.35 \\
FedCPA       &  {18.45$\pm$5.19} & 11.55$\pm$1.47  & 11.11$\pm$0.90 \\
FedMID       & 11.76$\pm$1.82  & 11.68$\pm$1.28  & 11.10$\pm$0.98  \\ \hline
\end{tabular}
\label{tab:appendix_model_size_tarasr}
\end{table}
%
\begin{table}[!ht]
\caption{Results varying the backbone network in targeted attacks. ACC is reported.}
\centering
%
\begin{tabular}{l|lll}
\hline
\multicolumn{1}{c|}{\multirow{2}{*}{Targeted (ACC)}} & \multicolumn{3}{c}{Model Size}                                                              \\ \cline{2-4} 
\multicolumn{1}{c|}{}                              & \multicolumn{1}{c}{Resnet18} & \multicolumn{1}{c}{Resnet50} & \multicolumn{1}{c}{Resnet101} \\ \hline
FedAvg   & 69.31$\pm$3.74 & 83.38$\pm$1.75  & 84.59$\pm$1.35  \\
Median       & 62.37$\pm$3.32 & 80.07$\pm$2.53  & 80.84$\pm$1.56  \\
Trimmed Mean & 71.42$\pm$2.77 & 84.88$\pm$1.06  & 84.65$\pm$1.12  \\
Multi-Krum   & 68.97$\pm$2.21 & 85.16$\pm$1.38  & 85.27$\pm$1.56  \\
FoolsGold    & 49.02$\pm$9.46 & 66.12$\pm$12.40 & 83.92$\pm$1.10  \\
ResidualBase & 69.85$\pm$3.59 & 84.88$\pm$0.98  & 84.42$\pm$1.33  \\
RFA          & 70.06$\pm$3.37 & 83.07$\pm$2.01  & 68.68$\pm$17.86 \\
DnC          &  70.03$\pm$3.00	& 84.80$\pm$1.59	& 84.54$\pm$1.88     \\
Bucket          &   68.34$\pm$4.21	& 84.16$\pm$1.15 &	83.26$\pm$2.19   \\
FLTrust          &   69.91$\pm$2.11	& 70.08$\pm$0.74	& 58.45$\pm$1.43 \\
FedCPA       &  {68.68$\pm$4.18} & 80.37$\pm$2.01  & 83.24$\pm$1.10 \\
FedMID       & 74.32$\pm$2.64 & 85.86$\pm$1.55  & 84.78$\pm$1.63  \\ \hline
\end{tabular}
\label{tab:appendix_model_size_taracc}
\end{table}
\newpage
\noindent
\textbf{Varying the attacker ratio}
\begin{table}[!ht]
\caption{Results varying attacker ratio with untargeted attacks. ACC is reported.}
\centering
\begin{tabular}{l|llllll}
\hline
\multicolumn{1}{c|}{\multirow{2}{*}{Untargeted (ACC)}} & \multicolumn{4}{c}{Attacker Ratio}                                                                      \\ \cline{2-7} 
\multicolumn{1}{c|}{}                              &  \multicolumn{1}{c}{0 (clean)} &\multicolumn{1}{c}{0.1} & \multicolumn{1}{c}{0.15} & \multicolumn{1}{c}{0.2} & \multicolumn{1}{c}{0.25} &\\ \hline
FedAvg   & 75.14$\pm$2.92 & 71.81$\pm$2.16  & 72.31$\pm$2.53  & 67.82$\pm$3.49 & 66.15$\pm$4.97 \\
Median       & 69.40$\pm$4.23 &65.88$\pm$4.10  & 63.26$\pm$4.55  & 59.77$\pm$3.16 & 54.68$\pm$8.67 \\
Trimmed Mean & 74.73$\pm$3.21 & 70.87$\pm$2.63  & 70.18$\pm$3.12  & 71.43$\pm$3.47 & 64.09$\pm$5.88 \\
Multi-Krum   & 75.78$\pm$2.52 & 73.30$\pm$2.89  & 73.79$\pm$2.12  & 72.75$\pm$3.61 & 72.15$\pm$2.21 \\
FoolsGold    & 65.45$\pm$4.13 & 37.82$\pm$14.42 & 37.88$\pm$13.45 & 18.56$\pm$7.53 & 32.56$\pm$9.00 \\
ResidualBase & 74.33$\pm$2.85 & 72.29$\pm$2.45  & 72.83$\pm$2.48  & 73.61$\pm$3.40 & 66.88$\pm$4.74 \\
RFA          & 73.80$\pm$2.70 & 74.25$\pm$2.73  & 74.27$\pm$2.37  & 72.65$\pm$2.31 & 71.22$\pm$2.47 \\
DnC          & 75.90$\pm$2.13  &74.15$\pm$2.96	& 74.83$\pm$2.36	& 72.35$\pm$4.80	& 71.29$\pm$3.52     \\
Bucket          & 73.98$\pm$2.75 & 72.99$\pm$3.09	& 70.61$\pm$2.31	& 70.12$\pm$2.76	& 65.38$\pm$5.05  \\
FLTrust       & 75.16$\pm$2.11 & 69.24$\pm$2.82  & 68.66$\pm$2.68  & 65.61$\pm$2.44 & 63.26$\pm$4.13 \\
FedCPA       &70.79$\pm$2.65 & 74.18$\pm$1.88 &74.03$\pm$2.09 & 74.86$\pm$3.30 &74.61$\pm$2.79  \\
FedMID       & 74.02$\pm$2.62& 74.82$\pm$3.19 & 75.16$\pm$3.27  & 74.82$\pm$3.08  & 74.77$\pm$2.43 \\ \hline
\end{tabular}
\label{tab:appendix_attacker_ratio_unt}
\end{table}
%
\begin{table}[!ht]
\caption{Results varying attacker ratio with targeted attacks. ASR is reported.}
\centering
\begin{tabular}{l|llll}
\hline
\multicolumn{1}{c|}{\multirow{2}{*}{Targeted (ASR)}} & \multicolumn{4}{c}{Attacker Ratio}                                                                      \\ \cline{2-5} 
\multicolumn{1}{c|}{}                              & \multicolumn{1}{c}{0.1} & \multicolumn{1}{c}{0.15} & \multicolumn{1}{c}{0.2} & \multicolumn{1}{c}{0.25} \\ \hline
FedAvg   & 31.36$\pm$9.24  & 40.12$\pm$18.44 & 50.90$\pm$25.09 & 49.86$\pm$22.85 \\
Median       & 24.74$\pm$4.76  & 38.23$\pm$15.59 & 70.62$\pm$16.51 & 59.93$\pm$25.60 \\
Trimmed Mean & 35.98$\pm$11.11 & 45.42$\pm$21.61 & 18.99$\pm$10.29 & 59.56$\pm$28.53 \\
Multi-Krum   & 30.04$\pm$8.35  & 26.77$\pm$8.55  & 40.39$\pm$21.85 & 34.68$\pm$6.24  \\
FoolsGold    & 25.11$\pm$28.95 & 35.32$\pm$34.08 & 46.84$\pm$34.83 & 32.77$\pm$28.91 \\
ResidualBase & 31.81$\pm$10.40 & 51.98$\pm$20.79 & 54.00$\pm$27.50 & 59.48$\pm$27.37 \\
RFA          & 16.34$\pm$2.78  & 35.29$\pm$15.01 & 44.79$\pm$21.58 & 52.52$\pm$23.60 \\
DnC          &  24.45$\pm$13.66	& 36.30$\pm$17.26	& 36.03$\pm$16.60 &	48.61$\pm$28.27    \\
Bucket          &  32.11$\pm$10.56 &	44.11$\pm$14.34	& 50.91$\pm$21.97 &	48.87$\pm$28.66     \\
FLTrust          &  25.38$\pm$7.22 &	32.64$\pm$15.28	& 43.42$\pm$16.41 &	41.65$\pm$17.09     \\
FedCPA        &13.29$\pm$1.66 &14.32$\pm$1.93 & 18.45$\pm$5.19 &14.23$\pm$3.84  \\
FedMID       & 11.91$\pm$1.78  & 12.42$\pm$1.77  & 11.76$\pm$1.82  & 11.47$\pm$1.94  \\ \hline
\end{tabular}
\label{tab:appendix_attacker_ratio_tarasr}
\end{table}
%
\begin{table}[!ht]
\caption{Results varying attacker ratio with targeted attacks. ACC is reported.}
\centering
\begin{tabular}{l|llll}
\hline
\multicolumn{1}{c|}{\multirow{2}{*}{Targeted (ACC)}} & \multicolumn{4}{c}{Attacker Ratio}                                                                      \\ \cline{2-5} 
\multicolumn{1}{c|}{}                              & \multicolumn{1}{c}{0.1} & \multicolumn{1}{c}{0.15} & \multicolumn{1}{c}{0.2} & \multicolumn{1}{c}{0.25} \\ \hline
FedAvg   & 72.19$\pm$2.55  & 72.17$\pm$2.65 & 69.31$\pm$3.74 & 69.40$\pm$3.67 \\
Median       & 65.83$\pm$4.59  & 63.64$\pm$4.93 & 62.37$\pm$3.32 & 61.33$\pm$3.57 \\
Trimmed Mean & 71.81$\pm$3.11  & 71.80$\pm$2.99 & 71.42$\pm$2.77 & 68.51$\pm$3.42 \\
Multi-Krum   & 70.63$\pm$3.53  & 72.24$\pm$2.98 & 68.97$\pm$2.21 & 71.67$\pm$2.57 \\
FoolsGold    & 47.17$\pm$15.21 & 45.76$\pm$9.39 & 49.02$\pm$9.46 & 57.05$\pm$4.34 \\
ResidualBase & 73.09$\pm$2.70  & 73.36$\pm$2.77 & 69.85$\pm$3.59 & 69.59$\pm$3.06 \\
RFA          & 73.45$\pm$2.41  & 72.40$\pm$2.43 & 70.06$\pm$3.37 & 68.83$\pm$4.59 \\
DnC          & 72.50$\pm$2.57 &	72.67$\pm$2.91	& 70.03$\pm$3.00	& 69.09$\pm$5.51             \\
Bucket          &  72.99$\pm$3.09 &	71.99$\pm$2.73 &	68.34$\pm$4.21 &	69.30$\pm$3.42       \\
FLTrust          &  71.31$\pm$2.23 &	70.13$\pm$2.83	& 69.91$\pm$2.11 &	68.74$\pm$2.34     \\
FedCPA    &72.59$\pm$2.70 &72.41$\pm$2.64 & 68.68 $\pm$4.18 &66.70$\pm$5.21  \\
FedMID       & 74.57$\pm$2.83  & 75.81$\pm$3.10 & 74.32$\pm$2.64 & 74.11$\pm$3.35 \\ \hline
\end{tabular}
\label{tab:appendix_attacker_ratio_taracc}
\end{table}
\newpage
\noindent
\textbf{Varying the number of clients}
\begin{table}[!ht]
\caption{Results varying the number of local clients with untargeted attacks. ACC is reported.}
\centering
\begin{tabular}{l|llll}
\hline
\multicolumn{1}{c|}{\multirow{2}{*}{Untargeted (ACC)}} & \multicolumn{4}{c}{The number of local clients}                                                  \\ \cline{2-5} 
\multicolumn{1}{c|}{}                              & \multicolumn{1}{c}{5} & \multicolumn{1}{c}{10} & \multicolumn{1}{c}{20} & \multicolumn{1}{c}{30} \\ \hline
FedAvg   & 87.35$\pm$1.05  & 82.67$\pm$0.65  & 67.82$\pm$3.49 & 55.77$\pm$4.59  \\
Median       & 85.36$\pm$0.96  & 81.18$\pm$0.73  & 59.77$\pm$3.16 & 46.81$\pm$6.42  \\
Trimmed Mean & 85.76$\pm$0.73  & 82.40$\pm$0.63  & 71.43$\pm$3.47 & 55.07$\pm$4.70  \\
Multi-Krum   & 88.67$\pm$0.35  & 84.06$\pm$0.64  & 72.75$\pm$3.61 & 58.10$\pm$4.94  \\
FoolsGold    & 30.74$\pm$21.34 & 53.48$\pm$18.38 & 18.56$\pm$7.53 & 41.10$\pm$11.11 \\
ResidualBase & 88.91$\pm$0.25  & 83.53$\pm$0.69  & 73.61$\pm$3.40 & 56.27$\pm$4.22  \\
RFA          & 87.33$\pm$0.38  & 82.74$\pm$0.83  & 72.65$\pm$2.31 & 58.17$\pm$5.20  \\
DnC          & 89.01$\pm$0.29	& 83.28$\pm$0.81	& 72.35$\pm$4.80 &	52.12$\pm$4.19   \\
Bucket          & 86.89$\pm$1.53	& 80.54$\pm$1.42	& 70.12$\pm$2.76	& 51.99$\pm$6.68 \\
FLTrust          & 84.13$\pm$1.35	& 73.33$\pm$1.01	& 65.61$\pm$2.44	& 54.17$\pm$3.39 \\
FedCPA          & 88.85$\pm$0.40	& 85.80$\pm$0.39	& 74.86$\pm$3.30	& 58.12$\pm$2.72 \\
FedMID       & 89.04$\pm$0.28  & 83.97$\pm$0.34  & 74.82$\pm$0.38 & 59.56$\pm$3.17  \\ \hline
\end{tabular}
\label{tab:appendix_local_clients_unt}
\end{table}
\begin{table}[!ht]
\caption{Results varying the number of local clients with targeted attacks. ASR is reported.}
\centering
\begin{tabular}{l|llll}
\hline
\multicolumn{1}{c|}{\multirow{2}{*}{Targeted (ASR)}} & \multicolumn{4}{c}{The number of local clients}                                                  \\ \cline{2-5} 
\multicolumn{1}{c|}{}                              & \multicolumn{1}{c}{5} & \multicolumn{1}{c}{10} & \multicolumn{1}{c}{20} & \multicolumn{1}{c}{30} \\ \hline
FedAvg   & 53.33$\pm$7.25  & 64.07$\pm$4.80 & 50.90$\pm$25.09 & 52.10$\pm$29.63 \\
Median       & 64.67$\pm$11.85 & 63.18$\pm$6.12 & 70.62$\pm$16.51 & 59.47$\pm$16.25 \\
Trimmed Mean & 61.55$\pm$11.86 & 62.51$\pm$4.95 & 18.99$\pm$10.29 & 47.97$\pm$30.20 \\
Multi-Krum   & 54.54$\pm$14.46 & 55.25$\pm$6.28 & 40.39$\pm$21.85 & 40.87$\pm$28.46 \\
FoolsGold    & 86.75$\pm$17.97 & 39.31$\pm$7.98 & 46.84$\pm$34.83 & 46.51$\pm$31.79 \\
ResidualBase & 55.57$\pm$11.76 & 62.67$\pm$5.03 & 54.00$\pm$27.50 & 52.30$\pm$29.89 \\
RFA          & 52.39$\pm$2.79  & 32.71$\pm$4.92 & 44.79$\pm$21.58 & 51.54$\pm$27.76 \\
DnC          & 19.59$\pm$14.13 &	15.42$\pm$2.95 &	36.03$\pm$16.60	& 48.92$\pm$36.98          \\
Bucket          &  55.28$\pm$15.62	& 52.84$\pm$28.42	& 50.91$\pm$21.97	& 52.04$\pm$28.63         \\
FLTrust          & 45.18$\pm$4.14	& 57.08$\pm$4.43	& 43.42$\pm$16.41	& 43.30$\pm$25.87 \\
FedCPA          & 10.46$\pm$0.37	& 10.87$\pm$0.29	& 18.45$\pm$5.19	& 25.13$\pm$4.33 \\
FedMID       & 10.19$\pm$0.27  & 11.07$\pm$0.49 & 11.76$\pm$1.82  & 29.31$\pm$17.26 \\ \hline
\end{tabular}
\label{tab:appendix_local_clients_tarasr}
\end{table}
%
\begin{table}[!ht]
\caption{Results varying the number of local clients with targeted attacks. ACC is reported.}
\centering
\begin{tabular}{l|llll}
\hline
\multicolumn{1}{c|}{\multirow{2}{*}{Targeted (ASR)}} & \multicolumn{4}{c}{The number of local clients}                                                  \\ \cline{2-5} 
\multicolumn{1}{c|}{}                              & \multicolumn{1}{c}{5} & \multicolumn{1}{c}{10} & \multicolumn{1}{c}{20} & \multicolumn{1}{c}{30} \\ \hline
FedAvg   & 88.92$\pm$0.22 & 84.39$\pm$0.21 & 69.31$\pm$3.74 & 55.24$\pm$7.76  \\
Median       & 87.62$\pm$0.36 & 81.43$\pm$0.38 & 62.37$\pm$3.32 & 48.47$\pm$5.53  \\
Trimmed Mean & 88.38$\pm$0.36 & 83.72$\pm$0.30 & 71.42$\pm$2.77 & 54.48$\pm$7.96  \\
Multi-Krum   & 87.45$\pm$0.35 & 83.25$\pm$0.32 & 68.97$\pm$2.21 & 54.70$\pm$6.61  \\
FoolsGold    & 78.30$\pm$2.57 & 74.41$\pm$1.74 & 49.02$\pm$9.46 & 39.02$\pm$12.60 \\
ResidualBase & 89.03$\pm$0.25 & 83.66$\pm$0.36 & 69.85$\pm$3.59 & 55.82$\pm$6.79  \\
RFA          & 85.51$\pm$0.50 & 82.92$\pm$0.55 & 70.06$\pm$3.37 & 58.96$\pm$6.55  \\
DnC          &   88.56$\pm$0.77 &	83.74$\pm$0.68	& 70.03$\pm$3.00	& 53.08$\pm$5.40          \\
Bucket          &   86.80$\pm$1.15	& 82.07$\pm$0.94	& 68.34$\pm$4.21	& 54.49$\pm$7.56      \\
FLTrust          & 89.32$\pm$0.44	& 83.80$\pm$0.28	& 69.91$\pm$2.11	& 58.68$\pm$1.84 \\
FedCPA          & 89.20$\pm$0.27	& 82.54$\pm$0.64	& 68.68$\pm$4.18	& 55.49$\pm$5.04 \\
FedMID       & 88.81$\pm$0.57 & 84.25$\pm$0.53 & 74.32$\pm$2.64 & 60.12$\pm$3.40  \\ \hline
\end{tabular}
\label{tab:appendix_local_clients_taracc}
\end{table}
\newpage

\noindent
\textbf{Varying the base federated learning algorithm - FedProx} \\
\noindent
\suw{To ensure the robustness of our proposed model under varied conditions, we replicated the Scenario-1 experiment using an alternative federated learning algorithm. Specifically, we employed FedProx~\cite{li2020fedprox} in place of FedAvg as the foundational federated learning approach, which is often used to deal with heterogeneity in federated systems.} 
\begin{table*}[!ht]
\caption{Experimental results under untargeted label flipping attacks with FedProx. ACC is reported.}
\centering
\begin{tabular}{l|cccc}
\hline
\multicolumn{1}{c|}{\begin{tabular}[c]{@{}c@{}}Untargeted (ACC)\end{tabular}} &
  CIFAR10 &
  CIFAR100 &
  SVHN &
  \begin{tabular}[c]{@{}c@{}} TinyImageNet \end{tabular} \\ \hline
FedAvg   & 68.98$\pm$2.76& 42.8$\pm$1.25& 88.86$\pm$2.39& 34.08$\pm$4.09\\
Median       & 59.03$\pm$5.06& 36.1$\pm$1.73& 89.56$\pm$1.69& 28.96$\pm$3.63 \\
Trimmed Mean & 66.82$\pm$3.29& 43.6$\pm$1.17& 90.14$\pm$1.68& 35.07$\pm$3.35 \\
Multi-Krum   & 73.72$\pm$2.74& 47.6$\pm$0.88& 92.34$\pm$0.77& 36.93$\pm$1.66 \\
FoolsGold    & 32.34$\pm$13.26& 8.7$\pm$6.71& 61.45$\pm$19.82& 5.58$\pm$4.75 \\
ResidualBase & 69.87$\pm$2.64& 42.8$\pm$3.45& 91.88$\pm$1.06& 36.36$\pm$2.63 \\
RFA          & 73.04$\pm$2.37& 48.2$\pm$0.82& 92.30$\pm$1.08& 36.09$\pm$0.96 \\
DnC          &  73.39$\pm$2.36& 46.4$\pm$0.98& 92.16$\pm$0.88& 36.57$\pm$2.31 \\        
Bucket          & 71.07$\pm$2.90& 45.8$\pm$0.67& 90.68$\pm$2.83& 37.04$\pm$1.60 \\
FLTrust & 66.85$\pm$2.73& 41.5$\pm$1.11& 88.19$\pm$1.32& 33.56$\pm$2.28 \\
FedCPA & 73.63$\pm$3.66& 42.7$\pm$1.91& 92.81$\pm$0.94& 37.50$\pm$1.27 \\
FedMID & 75.91$\pm$3.03& 48.2$\pm$0.38& 92.89$\pm$0.54& 36.47$\pm$0.76 \\ \hline
\end{tabular}
\label{tab:appendix_unt_acc_fedprox}
\end{table*}
\begin{table*}[!ht]
\caption{Experimental results under targeted label flipping attacks with FedProx. ASR is reported.}
\centering
\begin{tabular}{l|cccc}
\hline
\multicolumn{1}{c|}{\begin{tabular}[c]{@{}c@{}}Targeted (ASR)\end{tabular}} &
  CIFAR10 &
  CIFAR100 &
  SVHN &
  \begin{tabular}[c]{@{}c@{}} TinyImageNet \end{tabular} \\ \hline
FedAvg   & 51.46$\pm$20.67& 45.06$\pm$39.90& 21.55$\pm$1.65& 96.74$\pm$0.53 \\
Median       & 47.79$\pm$21.34& 11.36$\pm$11.92& 22.21$\pm$1.87& 96.03$\pm$0.69 \\
Trimmed Mean & 53.97$\pm$25.49& 48.71$\pm$37.76& 21.67$\pm$1.71& 97.03$\pm$0.77 \\
Multi-Krum   & 31.22$\pm$8.46& 0.98$\pm$1.61& 21.77$\pm$2.15& 11.98$\pm$5.56 \\
FoolsGold    & 43.94$\pm$21.65& 58.80$\pm$43.15& 28.26$\pm$18.18& 54.38$\pm$47.87 \\
ResidualBase & 58.21$\pm$21.09& 10.91$\pm$30.52& 21.46$\pm$1.35& 96.39$\pm$1.18 \\
RFA     & 45.79$\pm$10.07& 1.08$\pm$0.90& 20.83$\pm$1.51& 14.75$\pm$5.64 \\
DnC     &  30.69$\pm$18.52& 2.88$\pm$6.07& 20.94$\pm$1.43& 86.46$\pm$2.19 \\
Bucket  & 50.35$\pm$14.87& 7.65$\pm$6.05& 22.23$\pm$2.17& 74.77$\pm$38.03 \\
FLTrust & 41.42$\pm$13.64& 45.96$\pm$39.18& 21.76$\pm$0.83& 96.53$\pm$0.71 \\
FedCPA & 18.75$\pm$5.44& 0.72$\pm$0.51& 20.51$\pm$1.05& 0.75$\pm$0.23 \\
FedMID & 12.01$\pm$1.93& 0.58$\pm$0.29& 21.17$\pm$1.06& 0.68$\pm$0.31 \\ \hline
\end{tabular}
\label{tab:appendix_targeted_asr_fedprox}
\end{table*}
\begin{table*}[!ht]
\caption{Experimental results under targeted label flipping attacks with FedProx. ACC is reported.}
\centering
\begin{tabular}{l|cccc}
\hline
\multicolumn{1}{c|}{Targeted (ACC)} & CIFAR10 & CIFAR100 & SVHN & TinyImageNet\\ \hline
FedAvg   & 70.68$\pm$2.67& 37.76$\pm$5.18& 91.50$\pm$0.78& 39.14$\pm$0.35 \\
Median       & 62.14$\pm$5.01& 34.83$\pm$1.95& 89.84$\pm$1.16& 31.54$\pm$0.72 \\
Trimmed Mean & 69.41$\pm$3.43& 38.91$\pm$4.36& 91.59$\pm$0.85& 38.35$\pm$0.39 \\
Multi-Krum   & 69.82$\pm$3.71& 47.50$\pm$0.84& 90.59$\pm$1.43& 36.98$\pm$0.95 \\
FoolsGold    & 60.80$\pm$2.73& 6.88$\pm$5.31& 70.56$\pm$18.62& 15.79$\pm$4.19 \\
ResidualBase & 71.21$\pm$2.32& 39.56$\pm$7.01& 91.96$\pm$0.75& 37.81$\pm$0.42 \\
RFA          & 72.44$\pm$2.22& 47.41$\pm$0.92& 92.27$\pm$0.92& 36.94$\pm$0.92 \\
DnC          &   70.96$\pm$2.53& 45.72$\pm$1.57& 92.66$\pm$0.64& 37.08$\pm$0.91 \\
Bucket          &  70.15$\pm$2.93& 45.09$\pm$0.71& 90.05$\pm$1.85& 36.60$\pm$4.12 \\
FLTrust       &  68.78$\pm$2.28& 36.78$\pm$5.26& 91.59$\pm$0.65& 38.63$\pm$0.38 \\
FedCPA&       69.52$\pm$3.37& 40.27$\pm$1.14& 93.43$\pm$0.39& 37.79$\pm$0.76 \\
FedMID       &  74.16$\pm$2.66& 47.31$\pm$1.01& 93.11$\pm$0.50& 37.01$\pm$0.68 \\ \hline
\end{tabular}
\label{tab:appendix_tar_acc_fedprox}
\end{table*}

\newpage

\section{Appendix C - Computational Complexity}
All experiments were deployed on four A100 GPUs. A comparative analysis of the time costs for each round of training across different defense strategies is provided in Table~\ref{Tab:appendix_complexity}, using a scenario with 20 clients and the CIFAR-10 dataset. Our proposed model, \model{}, required approximately 25\% more processing time compared to the conventional FedAvg algorithm.
\begin{table}[!ht]
\caption{Comparative analysis of relative time costs among defense strategies.}
\centering
\scalebox{1.1}{
\begin{tabular}{l|ccc}
\toprule
Method       & Relative Time Cost
 \\ \midrule
FedAvg & 1.00 \\
Median & 1.27 \\
Trimmed Mean & 1.03 \\
Multi-Krum & 1.06 \\
FoolsGold & 1.04 \\
ResidualBase & 1.99 \\
RFA & 1.00 \\
DnC & 1.06 \\
Bucket & 1.03 \\
FLTrust & 1.04 \\
FedCPA & 1.39\\
FedMID & 1.22 \\ \bottomrule
\end{tabular}}
\label{Tab:appendix_complexity}
\end{table}
\section{Appendix D - Qualitative Analysis}
\begin{figure}[!ht]
    \centering
    \includegraphics[height=8cm]{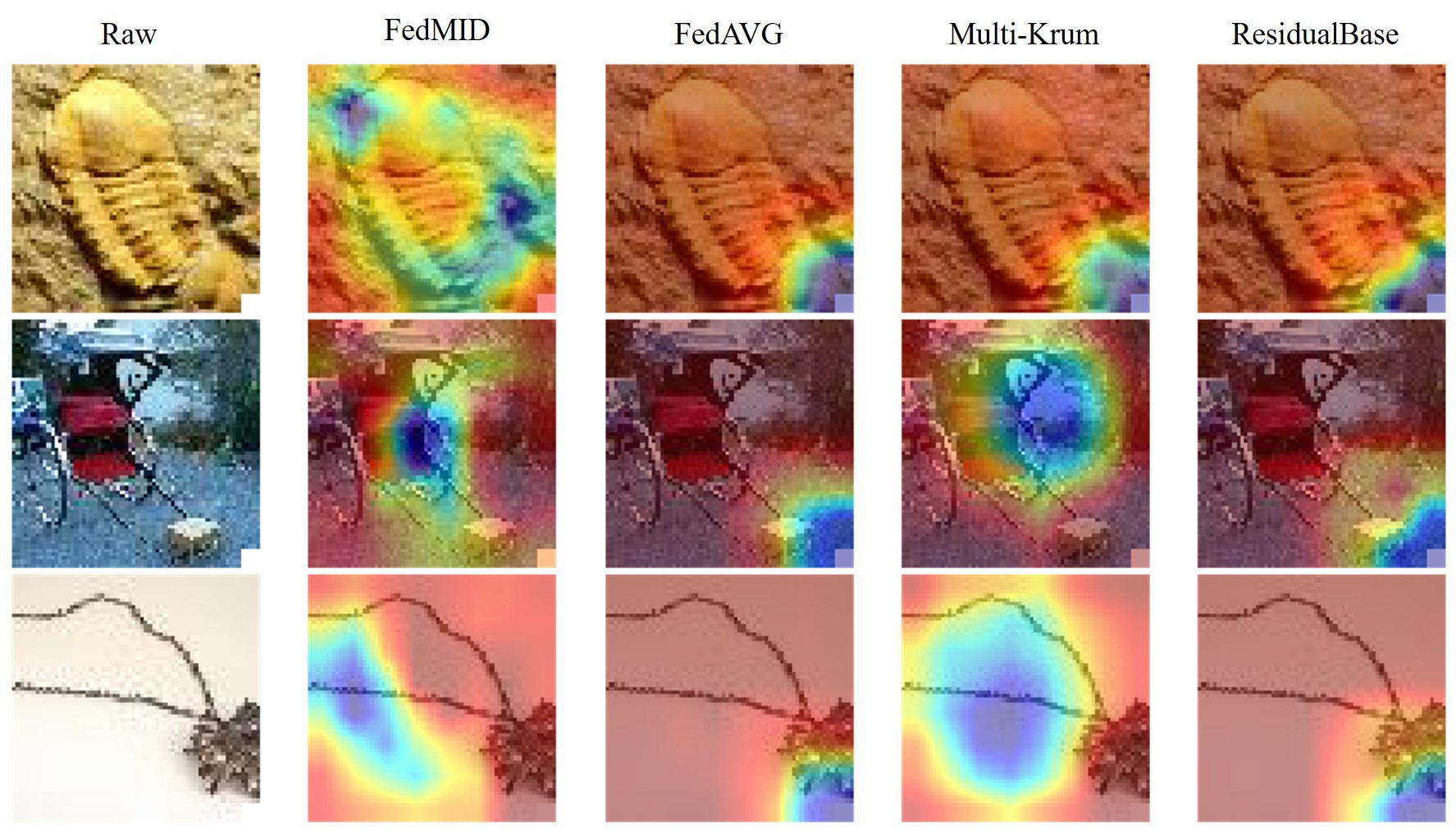}
    \caption{Grad-CAM visualization of global model's prediction from different defense strategies under targeted attacks over TinyImageNet.
    }
    \label{fig:qualitative_analysis}    
\end{figure}

We conducted a qualitative analysis to evaluate the effectiveness of \model{} in filtering out malicious knowledge during training in targeted attack scenarios.
We compared the performance of various defense strategies in understanding class features after training, as illustrated in Figure~\ref{fig:qualitative_analysis}.
To assess the interpretability of each model, we corrupted test set images with a small noise patch employed by attackers and utilized the Grad-CAM algorithm~\cite{selvaraju2017grad} to visualize the model's focus on each input.
In contrast to other scenarios where the model picks up malicious knowledge and concentrates on the inserted noise patch, our method typically extracts important features from the image. \looseness=-1

\end{document}